\documentclass[twoside,11pt]{article}

\usepackage[preprint]{jmlr_arxiv}

\usepackage[utf8]{inputenc} 
\usepackage[T1]{fontenc}    
\usepackage{hyperref}       
\usepackage{url}            
\usepackage{booktabs}       
\usepackage{amsfonts}       
\usepackage{nicefrac}       
\usepackage{microtype}      
\usepackage{xcolor}         

\usepackage{amsmath} 
\usepackage{graphicx} 
\usepackage{bm}
\usepackage{subcaption}
\usepackage{cancel}

\usepackage{lastpage}
\jmlrheading{TBC}{TBC}{TBC-\pageref{LastPage}}{2/24; Revised TBC}{TBC}{21-0000}{Thomas Baldwin-McDonald and Mauricio A. \'Alvarez}

\ShortHeadings{Deep LFMs: ODE-based Process Convolutions for Bayesian Deep Learning}{Baldwin-McDonald and \'Alvarez}
\firstpageno{1}

\begin{document}

\title{Deep Latent Force Models: ODE-based Process Convolutions for Bayesian Deep Learning}

\author{\name Thomas Baldwin-McDonald \email thomas.mcdonald-2@postgrad.manchester.ac.uk \\
    \name Mauricio A. \'Alvarez \email mauricio.alvarezlopez@manchester.ac.uk \\
       \addr Department of Computer Science \\ University of Manchester \\ Manchester \\ United Kingdom}
       
\editor{TBC}

\maketitle

\begin{abstract}
Modelling the behaviour of highly nonlinear dynamical systems with robust uncertainty quantification is a challenging task which typically requires approaches specifically designed to address the problem at hand. We introduce a domain-agnostic model to address this issue termed the deep latent force model (DLFM), a deep Gaussian process with physics-informed kernels at each layer, derived from ordinary differential equations using the framework of process convolutions. Two distinct formulations of the DLFM are presented which utilise weight-space and variational inducing points-based Gaussian process approximations, both of which are amenable to doubly stochastic variational inference. We present empirical evidence of the capability of the DLFM to capture the dynamics present in highly nonlinear real-world multi-output time series data. Additionally, we find that the DLFM is capable of achieving comparable performance to a range of non-physics-informed probabilistic models on benchmark univariate regression tasks. We also empirically assess the negative impact of the inducing points framework on the extrapolation capabilities of LFM-based models.
\end{abstract}


\section{Introduction}
Across many different areas of scientific study, time-varying physical and biological processes are modelled as dynamical systems, with their behaviour described by a set of ordinary differential equations (ODEs). If the form of said system is known, and is sufficiently low-dimensional, we may attempt to model the behaviour of the system by inferring the parameters of its constituent ODEs using observational data \citep{meeds2019efficient, ghosh2021variational}. Unfortunately, in practice, real-world systems are often sufficiently complex that it is infeasible to characterise all of the processes contained within them, let alone the interactions between these processes. Instead of forming a \textit{fully} mechanistic model of a complex system, latent force models (LFMs) \citep{alvarez2009latent} construct a \textit{simplified} mechanistic model of the system in question, which allows us to capture the salient features of the dynamics of the system using a (typically small) number of \textit{latent forces}. This hybrid approach leverages the advantages of data-driven modelling, whilst retaining some vital advantages of mechanistic modelling, such as the ability to extrapolate outside of the training input domain.

\textit{Nonlinear} dynamical systems are generally more challenging to model, often containing features such as nonstationarities which shallow models typically struggle to capture and extrapolate effectively. Whilst there has been work on integrating nonlinear differential equations into LFMs \citep{alvarez2019non, ward2020black, ross2021learning}, an alternative approach to modeling nonlinear dynamics which we will consider in this work, is to utilise a deep model, formed using a composition of functions. Such architectures possess increased representational power relative to shallow models due to their hierarchical structure \citep{lecun2015deep}, and furthermore, \textit{Bayesian} deep models such as \textit{deep Gaussian processes} (DGPs) are able to leverage this representational power whilst also providing a robust quantification of the uncertainty in their outputs. In this paper, we consider the modelling of nonlinear dynamical systems using a DGP formed from compositions of ODE-informed kernels. This approach is motivated by two key factors; firstly, many real-world systems and processes can be represented as compositional hierarchies \citep{lecun2015deep}, and secondly, there is compelling evidence to suggest that shallow GP-based models are unable to optimally learn compositional functions \citep{giordano2022inability}.

We introduce a generalised framework termed the \textit{deep latent force model} (DLFM), a novel approach to formulating a physics-informed hierarchical probabilistic model which is summarised in Figure \ref{fig:dlfm_diagram_chain}. We present two approaches to formulating this class of model and performing computationally efficient approximate Bayesian inference. Some portions of this work have been previously published in \cite{mcdonald2021compositional}, therefore in the remainder of this section we will outline which portions of this paper have been previously published, and which portions are new contributions.

Firstly, in Section \ref{sec:dlfm_rff} we outline a random Fourier feature-based approach previously published in \citet{mcdonald2021compositional}. This involves the derivation of physics-informed random Fourier features, which we compute via the convolution of Fourier features corresponding to an exponentiated quadratic (EQ) latent GP prior, with the Green’s function associated with a first order ODE. These features are then incorporated into each layer of a DGP which uses weight-space approximations \citep{cutajar2017random}. In addition to the material previously published in \citet{mcdonald2021compositional}, in Sections \ref{sec:dlfm_rff_initial} and \ref{sec:dlfm_rff_reparam} we propose two new improvements to the original Fourier feature-based model, namely introducing learnable initial conditions and allowing the local reparameterization trick (used for gradient variance reduction during training) to be disabled at test time. However, the most significant additional contribution included in this paper but not present in \citet{mcdonald2021compositional} is the introduction in Section \ref{sec:dlfm_pathwise} of a novel formulation of the DLFM and associated inference scheme based on \textit{pathwise sampling} \cite{wilson2020efficiently} and inducing points, whereby we instead perform the aforementioned convolution integral using a closed form expression for samples from the latent EQ GP. To ensure the scalability of both schemes to large datasets, stochastic variational inference is employed as a method for approximate Bayesian inference.  

We provide experimental evidence that our modelling framework is capable of capturing highly nonlinear dynamics effectively in both toy examples and real world data, whilst also being applicable to more general regression problems. The experimental results provided in Section \ref{sec:experiments} have not been previously published in \citet{mcdonald2021compositional}, as they were obtained using the improved version of the Fourier feature-based model, as well as the new inducing points-based model, both of which are introduced in this paper. Finally, in Section \ref{sec:dlfm_discussion} we present new analysis and discussion surrounding the topic of how the DLFM can be interpreted, as well as the advantages and disadvantages of using both formulations of the model in different modelling scenarios.

\section{Background} \label{sec:background}
In this section, we introduce a number of concepts relevant to this work. We begin by discussing the formulation of latent force models, a class of physics-informed Gaussian processes which form the basis of our deep probabilistic model. Additionally, we discuss two different approaches to performing approximate Bayesian inference in GPs, deep GPs and LFMs. The first approach involves using random Fourier features and weight-space GP approximations to perform stochastic variational inference, and the second is another form of variational scheme which relies upon inducing points and pathwise sampling.

\subsection{Latent force models} \label{sec:background_lfm}
Latent force models (LFMs) \citep{alvarez2009latent} are GPs with physically-inspired kernel functions, which typically encode the behaviour described by a specific form of differential equation. Instead of taking a fully mechanistic approach and specifying all the interactions within a physical system, the kernel describes a simplified form of the system in which the behaviour is determined by $Q$ latent forces. Given an input data-point $t$ and a set of $D$ output variables $\{f_d(t)\}^D_{d=1}$, a LFM expresses each output as,
\begin{align} \label{eq:lfm}
f_d(t) = \sum_{q=1}^Q S_{d, q} \int_{0}^t G_d(t-\tau)u_q(\tau)d\tau,
\end{align}
where $G_d(\cdot)$ represents the Green's function for a certain form of linear differential equation, $u_q(t)\sim \mathcal{GP}(0, k_q(t,t'))$ represents the GP prior over the the $q$-th latent force, and $S_{d,q}$ is a sensitivity parameter weighting the influence of the $q$-th latent force on the $d$-th output. Due to the linear nature of the convolution used to compute $f_d(t)$, the outputs can also be described by a GP. The general expression for the covariance of a LFM is given by $k_{f_d, f_{d'}}(t,t') = \sum_{q=1}^Q S_{d,q} S_{d',q} \int^t_0 G_d(t-\tau) \int^{t'}_0 G_{d'}(t'-\tau') k_q(\tau, \tau') d\tau' d\tau$. 

LFMs fall within the \textit{process convolution} framework for constructing GP kernels, which generally involves applying a convolution operator to a simple base GP in order to obtain an expressive covariance function \citep{alvarez2011computationally}. In the case of LFMs, the convolution filter has a fixed functional form $G_d(\cdot)$ with a physical interpretation, however, this need not be the case. For example, multiple works have instead placed a GP over the convolution filter \citep{mcdonald2022shallow, tobar2015learning}.

\subsection{Approximate inference with random Fourier features} \label{sec:background_rff}
Since their introduction to the machine learning research community by \citet{rahimi2007random}, \textit{random Fourier features} (RFFs) have been widely applied as a means of scaling kernel methods. Here we discuss their application to performing computationally efficient Bayesian inference in DGPs and LFMs, which we will leverage later in Section \ref{sec:dlfm_rff} in the context of our DLFM.

\subsubsection{RFFs for deep Gaussian processes}\label{sec:background_rff_dgp}
Gaussian processes (GPs) are nonparametric probabilistic models which
offer a great degree of flexibility in terms of the data they can
represent, however this flexibility is not unbounded. The hierarchical
networks of GPs now widely known as deep Gaussian processes (DGPs)
were first formalised by \citet{damianou2013deep}, with the motivating
factor behind their creation being the ability of deep networks to reuse features and allow for higher levels of abstraction in said features \citep{bengio2013representation}, which results in such models having more representational power and flexibility than shallow models such as GPs. DGPs
are effectively a composite function, where the input to the model is
transformed to the output by being passed through a series of latent
mappings (i.e. multivariate GPs). If we consider a supervised learning
problem with inputs denoted by $\mathbf{X} = \{\mathbf{x}_n\}^N_{n=1}$
and targets denoted by $\mathbf{y} = \{y_n\}^N_{n=1}$, we can write
the analytic form of the marginal likelihood for a DGP with $N_h$
hidden layers as,
\begin{align}
\begin{split}
p(\mathbf{y}|\mathbf{X}, \mathbf{\theta}) = \int
p(\mathbf{y}|\mathbf{F}^{(N_h)})p(\mathbf{F}^{(N_h)}|\mathbf{F}^{(N_h
  - 1)}, \mathbf{\theta}^{(N_h - 1)}) \times ... \times 
p(\mathbf{F}^{(1)}|\mathbf{X},
\mathbf{\theta}^{(0)})d\mathbf{F}^{(N_h)} \ ... \ d\mathbf{F}^{(1)},
\end{split}
\end{align}
where $\mathbf{F}^{(\ell)}$ and $\mathbf{\theta}^{(\ell)}$ represent
the latent values and covariance parameters respectively at the
$\ell$-th layer, where $\ell = 0, ... , N_h$. However, due to the
need to propagate densities through nonlinear GP covariance functions
within the model, this integral is intractable
\citep{damianou2015deep}. As this precludes us from employing exact
Bayesian inference in these models, various techniques for approximate inference have been applied to DGPs in recent years, with most approaches broadly based upon either variational inference \citep{salimbeni2017doubly, salimbeni2019deep, yu2019implicit} or Monte Carlo methods \citep{havasi2018inference}.

\citet{cutajar2017random} outline an alternative approach to tackling this
problem, which involves replacing the GPs present at each layer of the
network with their two layer weight-space approximation, 
forming a Bayesian neural network which acts as an approximation to a
DGP. Consider the $\ell$-th layer of such a DGP, which we assume is a zero-mean GP with an exponentiated quadratic kernel. Should this layer receive an input
$\mathbf{F}^{(\ell)}$, where $\mathbf{F}^{(0)} = \mathbf{X}$, the corresponding random features are denoted by 
$\mathbf{\Phi}^{(\ell)} \in \mathbb{R}^{N \times 2N^{(\ell)}_{RF}}$ and are given by,
\begin{align} \label{eq:dgp_features}
\mathbf{\Phi}^{(\ell)} = \sqrt[]{\frac{(\sigma^2)^{(\ell)}}{N_{RF}^{(\ell)}}} \left[ \cos(\mathbf{F}^{(\ell)}\mathbf{\Omega}^{(\ell)}), \sin(\mathbf{F}^{(\ell)}\mathbf{\Omega}^{(\ell)}) \right] ,
\end{align} 
where $(\sigma^2)^{(\ell)}$ is the marginal variance kernel hyperparameter, $N_{RF}^{(\ell)}$ is 
the number of random features used and $\mathbf{\Omega}^{(\ell)} \in \mathbb{R}^{D_{F^{(\ell)}} \times
  N_{RF}^{(\ell)}}$ is the matrix of spectral frequencies used to
determine the random features. This matrix is assigned a prior
$p(\Omega^{(\ell)}_d) = \mathcal{N}\left(\Omega^{(\ell)}_d \mid 0,
{(l^{(\ell)})}^{-2}\right)$ where $l^{(\ell)}$, is the lengthscale kernel hyperparameter,
$D_{F^{(\ell)}}$ is the number of GPs within the layer, and $d = 1, ..., D_{F^{(\ell)}}$. The
random features then undergo the linear transformation, $\mathbf{F}^{(\ell + 1)} = \mathbf{\Phi}^{(\ell)} \mathbf{W}^{(\ell)}$, where $\mathbf{W}^{(\ell)} \in \mathbb{R}^{2N_{RF}^{(\ell)} \times D_{F^{(\ell + 1)}}}$ is a weight matrix with each column assigned a standard normal prior. 
Training is achieved via \textit{stochastic variational inference}, which involves establishing a tractable lower bound for the marginal likelihood and optimizing said bound with respect to the mean and variance of the variational distributions over the weights and spectral frequencies across all layers of the network. The bound is also optimised with respect to the kernel hyperparameters across all layers.

\subsubsection{RFFs for latent force models}\label{sec:background_rff_lfm}
Exact inference in LFMs is tractable, however as with GPs, it scales with $\mathcal{O}(N^3)$ complexity \citep{Rasmussen06}. Typically, an EQ form is assumed for the kernel governing the latent forces, $k_q(\cdot)$, and by providing a random Fourier feature representation for this kernel, \citet{guarnizo2018fast} were able to reduce this cubic dependence on the number of data-points to a linear dependence. This representation arises from Bochner's theorem, which states,
\begin{align}
\begin{split}
k_q(\tau, \tau') &= e^{\frac{-(\tau-\tau')^2}{\ell_q^2}} 
=\int p(\omega)e^{j(\tau-\tau')\omega} d\omega \\ 
&\approx \frac{1}{N_{RF}}  \sum^{N_{RF}}_{s=1} e^{j\omega_s \tau} e^{-j\omega_s \tau'} , \label{eq:bochner2}
\end{split}
\end{align}
where $\ell_q$ is the lengthscale of the EQ kernel, $N_{RF}$ is the number of random features used in the approximation and $\omega_s \sim p(\omega) = \mathcal{N}(\omega | 0, 2/\ell_q^2)$. Substituting this form of the EQ kernel into the LFM covariance $k_{f_d, f_{d'}}(t,t')$ leads to a fast approximation given by,
\begin{align}
\begin{split}
\sum_{q=1}^Q \frac{S_{d,q}S_{d',q}}{N_{RF}} &\left[ \sum_{s=1}^{N_{RF}} \phi_d(t, \theta_d, \omega_s) \phi_{d'}^*(t', \theta_{d'}, \omega_s) \right] \\
\text{where,} \quad \phi_d(t, \theta_d, \omega) &= \int_0^t G_d(t-\tau)e^{j\omega \tau} d\tau, \label{eq:RFRF}
\end{split}
\end{align}
with $\theta_d$ representing the Green's function parameters. When the Green's function is a real function, $\phi_{d'}^*(t', \theta_{d'}, \omega) = \phi_{d'}(t', \theta_{d'}, -\omega)$. \citet{guarnizo2018fast} refer to $\phi_d(t, \theta_d, \omega)$ as \emph{random Fourier response features} (RFRFs).

\subsection{Approximate inference with pathwise sampling} \label{sec:background_pathwise}
We have seen that RFF approximations allow us to perform computationally efficient approximate Bayesian inference in both DGPs, and GPs in the process convolution framework (e.g. LFMs). However, a more widely used technique which also fulfills both of these criteria is the variational inducing points framework for approximate GP inference \citep{titsias2009variational}. In the case of a single output regression problem for a shallow GP, this involves introducing a set of inducing inputs $\mathbf{z} = \{\mathbf{z}_i \}_{i=1}^M$ and associated inducing variables $\mathbf{u} \in \mathbb{R}^{M}$. We can then parameterise these variables using a distribution of the form $q(\mathbf{u}) = \mathcal{N}(\mathbf{u} \mid \mathbf{m}, \mathbf{S})$, and optimise $\mathbf{m}$ and $\mathbf{S}$ by maximizing the aforementioned variational lower bound to the marginal likelihood \citep{hensman2013gaussian}. As well as being pervasive throughout the GP literature, variants of this approach also form the dominant paradigm for performing inference in DGPs \citep{salimbeni2017doubly, salimbeni2019deep, blomqvist2020deep, yu2019implicit}.

Performing any flavour of variational inference with GPs is predicated on the ability to obtain samples from the posterior in a computationally efficient manner, and this is also the case within the inducing points framework. \citet{wilson2020efficiently} introduced a novel approach to doing so based on Matheron's rule, which allows samples to evaluated at $N$ locations with $O(N)$ time complexity, which is a significant improvement over the typical $O(N^3)$ scaling associated with GPs. Another key benefit of their approach, which we will exploit in this work, is that the authors obtain \textit{functional} samples from the GP, which allows us to apply integral and differential operators directly to the samples, in order to efficiently generate samples from complex, non-Gaussian processes. This general approach was first introduced by \citet{ross2021learning} in order to generate samples from the output of a nonlinear process convolution, and has also been utilised by \citet{mcdonald2023nonparametric} in the context of building more expressive GP covariances.

Given a set of inducing variables $\mathbf{u}$ (or alternatively, data), the expression for functional samples from a GP posterior presented by \citet{wilson2020efficiently} takes the following form,
\begin{align}\label{eq:functional_samples}
    \underbrace{(f \mid \mathbf{u})(\cdot)}_{\text {posterior }} \stackrel{\mathrm{d}}{=} \underbrace{f(\cdot)}_{\text {prior }}+\underbrace{k(\cdot, \mathbf{Z}) \mathbf{K}^{-1}\left(\mathbf{u}-\mathbf{f}\right)}_{\text {update }}.
\end{align}
Here, $\mathbf{K}$ denotes the covariance matrix of the inducing variables with corresponding inputs $\mathbf{Z}$, and $\mathbf{f}=f(\mathbf{Z})$. This expression shows that we can decompose functional samples from the posterior (conditioned on data) into functional samples from the prior, and an update term which quantifies the residual between the prior sample and the data. \citet{wilson2020efficiently} employ a RFF representation for the prior, which reduces the computational burden of sampling considerably. However, the key insight from this work is that the pathologies associated with using RFFs in a nonstationary posterior are avoided, since only the prior (which has a stationary covariance) uses RFFs.

In Section \ref{sec:dlfm_pathwise}, we will show how we can use this inference scheme in the context of an LFM layer, allowing us to construct a deep LFM within the variational inducing points framework.

\section{Deep Latent Force Models} \label{sec:dlfm}
\begin{figure} 
    \centering
  \subfloat[Deep Gaussian process (DGP) \label{fig:dlfm_diagram_chain_a}]{%
       \includegraphics[width=0.9\linewidth]{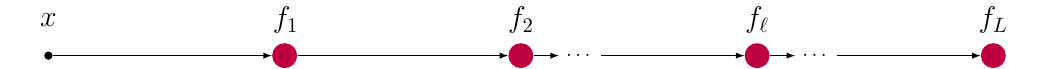}}
    \hfill
  \subfloat[Deep latent force model (DLFM) \label{fig:dlfm_diagram_chain_b}]{%
        \includegraphics[width=0.9\linewidth]{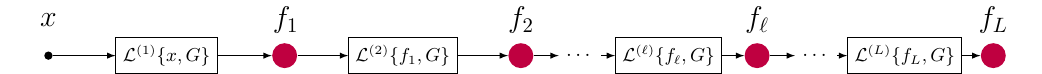}}
  \caption{A conceptual explanation of how the DLFM differs from a DGP. At each layer, we perform the operation $\mathcal{L}^{(\ell)}\{x, G\} = \int_0^x G^{(\ell)}(x-\tau)u(\tau)d\tau$, where $G$ is the Green's function corresponding to an ODE, and $u(\cdot)$ represents an exponentiated quadratic GP prior. For example, the second operation in the model shown above would take the form, $\mathcal{L}^{(2)}\{f_1, G\} = \int_0^{f_1} G^{(2)}(f_1-\tau)u(\tau)d\tau$.}
  \label{fig:dlfm_diagram_chain} 
\end{figure}
\textit{Deep latent force models} (DLFMs) are compositions of the LFMs described in Section \ref{sec:background_lfm}, in much the same way that DGPs are compositions of GPs. This relationship is shown in Fig. \ref{fig:dlfm_diagram_chain}, which conveys the fact that DLFMs are DGPs, but with an additional convolution operation performed at each layer of the compositional hierarchy. This convolution allows us to encode the dynamics of physical systems within the covariance of the GP layer, using the Green's function corresponding to a certain form of differential equation. 

In this section, we will consider two distinct approaches to formulating a DLFM. Firstly, in Section \ref{sec:dlfm_rff}, we will revisit the approach previously presented in \citet{mcdonald2021compositional} which utilises the weight-space formulation of GPs and random Fourier feature approximations. In addition to reviewing this previously published work we also introduce two modifications which improve the performance of the model, namely freely optimizable initial conditions and an option to disable local reparameterization at test time. Following this, in Section \ref{sec:dlfm_pathwise} we propose a new approach which uses an inducing points-based framework reliant on pathwise sampling. This new formulation of the DLFM is designed to address the issues associated with using a Fourier basis to represent a nonstationary posterior, which will be discussed in further detail. Our method builds upon prior work by \citet{mcdonald2023nonparametric}, and involves drawing functional samples from a latent GP and mapping them analytically through our convolution integral.

\subsection{Deep LFMs with RFFs} \label{sec:dlfm_rff}
\begin{figure} 
    \centering
  \subfloat[Deep Gaussian process (DGP) \label{fig:dlfm_diagram_2a}]{%
       \includegraphics[width=0.9\linewidth]{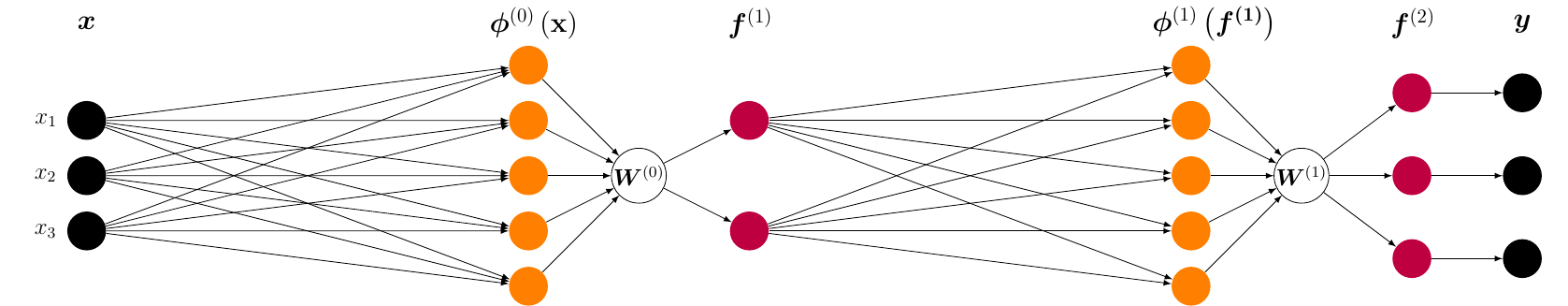}}
    \hfill
  \subfloat[Deep latent force model (DLFM-RFF) \label{fig:dlfm_diagram_2b}]{%
        \includegraphics[width=0.9\linewidth]{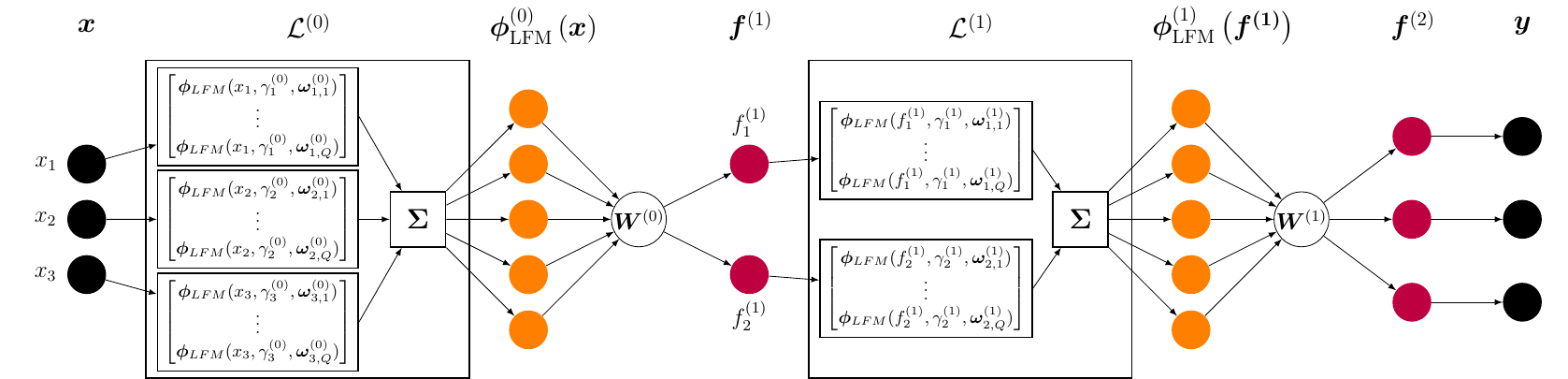}}
  \caption{An illustration of how the DLFM with random Fourier features differs from a DGP with random feature expansions, with this example containing two layers. At each layer of the DLFM-RFF, for each input dimension, $N_{RF}$ random features of the form shown in Eq. \eqref{eq:ode1_feature} are computed for each of the $Q$ latent forces. The random feature vector $\boldsymbol{\phi}_{LFM}^{(\ell)}$ is then formed by taking the sum of these features across the input dimensions. This summation is shown in the Figure by the block containing $\Sigma$.}
  \label{fig:dlfm_diagram_2} 
\end{figure}
Firstly we outline the variant of the DLFM previously presented in \citet{mcdonald2021compositional}, which utilises weight-space GP approximations and RFF-based variational inference. This model, which we will refer to as \textit{DLFM-RFF}, bears a number of similarities to the DGP introduced in Section \ref{sec:background_rff_dgp}. However, rather than deriving the random features $\mathbf{\Phi^{(\ell)}}$ within the DGP from an EQ kernel \citep{cutajar2017random}, we instead populate this matrix with features derived from an LFM kernel. The exact form of the features derived is dependent on the Green's function used, which in turn depends on the form of differential equation whose characteristics we wish to encode within the model. \citet{guarnizo2018fast} derived a number of different forms corresponding to various differential equations, with the simplest case being that of a first order ordinary differential equation (ODE) of the form,$\frac{df(t)}{dt} + \gamma f(t) = \sum^Q_{q=1} S_q u_q(t)$, where $\gamma$ is a decay parameter associated with the ODE and $S_q$ is a sensitivity parameter which weights the effect of each latent force. For simplicity of exposition, we assume $D=1$. The Green's function associated to this ODE has the form $G(x) = e^{-\gamma x}$. By using Eq. \eqref{eq:RFRF}, the RFRFs associated with this ODE follow as
\begin{align}
\phi(t, \gamma, \omega_s) & = \frac{e^{j\omega_s t} - e^{-\gamma t}}{\gamma + j\omega_s} ,  \label{eq:ode1_feature}
\end{align}
where compared to Eq. \eqref{eq:RFRF}, the parameter $\theta$ of the Green's function corresponds to the decay parameter $\gamma$.
From here on, we redefine the spectral frequencies as $\omega_{q, s}$ to emphasise the fact that the values sampled are dependent on the lengthscale of the 
latent force by way of the prior, $\omega_{q,s} \sim \mathcal{N}(\omega | 0, 2/\ell_q^2)$.

We can collect all of the random features corresponding to the $q$-th latent force into a single vector, $\boldsymbol{\phi}^c_q(t, \gamma, \boldsymbol{\omega}_q) = \sqrt{S_q^2/N_{RF}}[\phi(t,
  \gamma, \omega_{q, 1}), \cdots , \phi(t, \gamma, \omega_{q, N_{RF}})]^\top \in \mathbb{C}^{N_{RF} \times 1}$, where $\boldsymbol{\omega}_q=\{\omega_{q,s}\}_{s=1}^{N_{RF}}$, $\mathbb{C}$ refers to the complex plane and the super index $c$ in $\boldsymbol{\phi}_q^c(\cdot)$ makes explicit that this vector contains complex-valued numbers. By including the random features corresponding to all $Q$ latent forces within the model, we obtain $\boldsymbol{\phi}^c(t, \gamma, \boldsymbol{\omega}) = [\left(\boldsymbol{\phi}^c_1(t, \gamma, \boldsymbol{\omega}_1\right)^{\top}, \cdots , \left(\boldsymbol{\phi}^c_Q(t, \gamma, \boldsymbol{\omega}_Q)\right)^{\top}]^\top \in \mathbb{C}^{QN_{RF} \times 1}$, with $\boldsymbol{\omega}=\{\boldsymbol{\omega}_q\}_{q=1}^Q$. These random features will be denoted as $\boldsymbol{\phi}^c_{LFM}(t, \gamma, \boldsymbol{\omega})$ to differentiate them from the features $\boldsymbol{\phi}(\cdot)$ computed from a generic EQ kernel.

\subsubsection{Higher-dimensional inputs} Although the expression for $\boldsymbol{\phi}^c_{LFM}(t, \gamma,\boldsymbol{\omega})$ was obtained in the context of an ODE where the input is the time variable, we exploit this formalism to use these features even in the context of a generic supervised learning problem where the input is a potentially high-dimensional vector $\mathbf{x} = [x_1, x_2, \cdots, x_p]^\top\in\mathbb{R}^{p\times 1}$. As will be noticed later, such an extension is also necessary if we attempt to use such features at intermediate layers of the composition. Essentially, we compute a vector $\boldsymbol{\phi}^c_{LFM}(x_m, \gamma_m, \boldsymbol{\omega}_m)$ for each input dimension $x_{m}$ leading to a set of vectors  $\{\boldsymbol{\phi}^c_{LFM}(x_1, \gamma_1, \boldsymbol{\omega}_1),\cdots, \boldsymbol{\phi}^c_{LFM}(x_p, \gamma_p, \boldsymbol{\omega}_p)\}$. Notice that the samples $\boldsymbol{\omega}_m$ can also be different per input dimension, $x_m$. Although there are different ways in which these feature vectors can be combined, in this work, we assume that the final random feature vector computed over the whole input vector $\mathbf{x}$ is given as $\boldsymbol{\phi}^c_{LFM}(\mathbf{x}, \boldsymbol{\gamma}, \boldsymbol{\Omega}) = \sum_{m=1}^p \boldsymbol{\phi}^c_{LFM}(x_m, \gamma_m, \boldsymbol{\omega}_m)$, where $\boldsymbol{\Omega}=\{\boldsymbol{\omega}_m\}_{m=1}^p$ and $\boldsymbol{\gamma}=\{\gamma_m\}_{m=1}^p$. An alternative to explore for future work involves expressing $\boldsymbol{\phi}^c_{LFM}(\mathbf{x}, \boldsymbol{\gamma}, \boldsymbol{\Omega})$ as $\boldsymbol{\phi}^c_{LFM}(\mathbf{x}, \boldsymbol{\gamma}, \boldsymbol{\Omega}) = \sum_{m=1}^p \alpha_m \boldsymbol{\phi}^c_{LFM}(x_m, \gamma_m, \boldsymbol{\omega}_m)$, with $\alpha_m\in \mathbb{R}$ a parameter that weights the contribution of each input feature differently. Although we allow each input dimension to have a different decay parameter $\gamma_m$ in the experiments in Section \ref{sec:experiments}, for ease of notation we will assume that $\gamma_1=\gamma_2=\cdots=\gamma_p=\gamma$. For simplicity, we write $\boldsymbol{\phi}^c_{LFM}(\mathbf{x}, \gamma, \boldsymbol{\Omega})$. Therefore, $\boldsymbol{\phi}^c_{LFM}(\mathbf{x}, \gamma, \boldsymbol{\Omega})$ is a vector-valued function that maps from $\mathbb{R}^{p\times 1}$ to $\mathbb{C}^{QN_{RF} \times 1}$.

\subsubsection{Real version of the RFRFs} Rather than working with the complex-valued random features $\boldsymbol{\phi}^c_{LFM}(\mathbf{x}, \gamma, \boldsymbol{\Omega})$, we can work with their real-valued counterpart by using $\boldsymbol{\phi}_{LFM}(\mathbf{x}, \gamma, \boldsymbol{\Omega}) = [\left(\mathfrak{Re}\{\boldsymbol{\phi}^c_{LFM}(\mathbf{x}, \gamma, \boldsymbol{\Omega})\}\right)^\top,$ $\left(\mathfrak{Im}\{\boldsymbol{\phi}^c_{LFM}(\mathbf{x}, \gamma, \boldsymbol{\Omega})\}\right)^\top]^{\top}\in \mathbb{R}^{2QN_{RF}\times 1}$ \citep{guarnizo2018fast}, where $\mathfrak{Re}(a)$ and $\mathfrak{Im}(a)$
take the real component and imaginary component of $a$, respectively. For an input matrix $\mathbf{X} = [\mathbf{x}_1, \cdots, \mathbf{x}_N]^{\top}\in \mathbb{R}^{N\times p}$, the matrix $\boldsymbol{\Phi}_{LFM}(\mathbf{X}, \gamma, \boldsymbol{\Omega}) = [\boldsymbol{\phi}_{LFM}(\mathbf{x}_1, \gamma, \boldsymbol{\Omega}), \dots,$ $ \boldsymbol{\phi}_{LFM}(\mathbf{x}_N, \gamma, \boldsymbol{\Omega})]^{\top} \in \mathbb{R}^{N\times 2QN_{RF}}$.

\subsubsection{Composition of RFRFs} We now use $\boldsymbol{\Phi}_{LFM}(\mathbf{X}, \gamma, \boldsymbol{\Omega})$ as a building block of a layered architecture of RFRFs. We write $\boldsymbol{\Phi}^{(\ell)}_{LFM}(\mathbf{F}^{(\ell)}, \gamma^{(\ell)}, \boldsymbol{\Omega}^{(\ell)})\in \mathbb{R}^{N\times 2Q^{(\ell)}N^{(\ell)}_{RF}}$ and follow a similar construction to the one described in
Section \ref{sec:background_rff_dgp}, where $\mathbf{F}^{(\ell+1)} = \boldsymbol{\Phi}^{(\ell)}_{LFM}(\mathbf{F}^{(\ell)}, \gamma^{(\ell)}, \boldsymbol{\Omega}^{(\ell)})\mathbf{W}^{(\ell)}$, and $\mathbf{W}^{(\ell)}\in\mathbb{R}^{2Q^{(\ell)}N^{(\ell)}_{RF}\times D_{F^{(\ell+1)}}}$. As before, $\mathbf{F}^{(0)}=\mathbf{X}$. Fig. \ref{fig:dlfm_diagram_2} is an example of this compositional architecture of RFRFs, which we refer to as a \emph{deep latent force model} (DLFM). When considering multiple-output problems, we allow extra flexibility to the decay parameters and lengthscales at the final ($L$-th) layer such that they vary not only by input dimension, but also by output. Mathematically, this corresponds to computing $\mathbf{F}_d^{(L)} = \boldsymbol{\Phi}^{(L-1)}_{LFM}(\mathbf{F}^{(\ell)}, \gamma_d^{(L-1)}, \boldsymbol{\Omega}_d^{(L-1)})\mathbf{W}_d^{(L-1)}$, where $d=1,...,D$.

\subsubsection{Initial conditions} \label{sec:dlfm_rff_initial}
In this work, we introduce an additional component to the DLFM-RFF which was not previously studied in \citet{mcdonald2021compositional}, a consideration of the initial conditions of our system. The analytical Green's function we use for our first order ODE kernel is derived assuming an initial condition for the ODE of $y_d(t = 0) = 0$. This may seem simplistic, however given that we are assuming no a priori knowledge of the form of the dynamical systems we wish to model, it is challenging to reason about a more suitable alternative. The consequence of this is that samples from the model must obey this constraint, resulting in the model having limited ability to capture large variations in the output, in regions of input space close to the origin.

We partially address this by introducing a bias term $c_d$ for each output dimension within every layer of the DLFM-RFF, which we can interpret as a learnable initial condition of the first order dynamical system we study. Learning this parameter during training allows us greater flexibility with respect to the form of the functions we can represent, as we now have a scenario where $y_d(t = 0) = c_d$. However, we still have the restriction that all samples from the model must pass through this point.

In Section \ref{sec:dlfm_pathwise} we introduce another approach to addressing this in the context of our new formulation of the DLFM, which solves the problem and allows us to obtain samples which do not collapse to an initial condition at the origin.

\subsubsection{Reparameterization trick} \label{sec:dlfm_rff_reparam}
In \citet{mcdonald2021compositional} and \citet{cutajar2017random}, the \textit{local reparameterization trick} \citep{kingma2015variational} is employed in order to reduce the variance of the stochastic gradients computed when performing variational approximate Bayesian inference. Whilst this is desirable during training, it is not of concern at test time, and introduces noise into the predictions generated from the model. For this reason, we introduce a new option for the DLFM-RFF which allows reparameterization to be switched off at test time.

\subsubsection{Variational inference} \label{sec:dlfm_rff_vi}
As previously mentioned, exact Bayesian inference is intractable for models such as ours, therefore in order to train the DLFM we employ stochastic variational inference \citep{hoffman2013stochastic}. For notational simplicity, let $\mathbf{W}$, $\mathbf{\Omega}$ and $\mathbf{\Theta}$ represent the collections of weight matrices, spectral frequencies and kernel hyperparameters respectively, across all layers of the model. We seek to optimise the variational distributions over $\mathbf{\Omega}$ and $\mathbf{W}$ whilst also optimizing $\mathbf{\Theta}$, however we do not place a variational distribution over these hyperparameters. Our approach resembles the VAR-FIXED training strategy described by \citet{cutajar2017random} which involves reparameterizing $\Omega_{ij}^{(\ell)}$ such that $\Omega_{ij}^{(\ell)} = (s^2)_{ij}^{(\ell)} \epsilon_{rij}^{(\ell)} + m_{ij}^{(\ell)}$ , where $m_{ij}^{(\ell)}$ and $(s^2)_{ij}^{(\ell)}$ represent the means and variances associated with the variational distribution over $\Omega_{ij}^{(\ell)}$, and ensuring that the standard normal samples $\epsilon_{rij}^{(\ell)}$ are fixed throughout training rather than being resampled at each iteration. If we denote $\mathbf{\Psi} = \left\{\mathbf{W}, \mathbf{\Omega} \right\}$ and consider training inputs $\mathbf{X} \in \mathbb{R}^{N \times D_{in}}$ and outputs $\mathbf{y} \in \mathbb{R}^{N \times D_{out}}$, we can derive a tractable lower bound on the marginal likelihood using Jensen's inequality, which allows for minibatch training using a subset of $M$ observations from the training set of $N$ total observations. This bound, derived by \citet{cutajar2017random}, takes the form,
\begin{align}
\begin{split}
\log[p(\mathbf{y} | \mathbf{X}, \mathbf{\Theta}] &= E_{q(\mathbf{\Psi})} \log[p(\mathbf{y} | \mathbf{X, \Psi, \Theta})] - \text{D}_\text{KL}[q(\mathbf{\Psi}) || p(\mathbf{\Psi} | \mathbf{\Theta})] \\
&\approx \left[\frac{N}{M} \sum_{k \in \mathcal{I}_M} \frac{1}{N_{\text{MC}}} \sum^{N_{\text{MC}}}_{r=1} \log[p(\mathbf{y}_k | \mathbf{x}_k, \tilde{\mathbf{\Psi}}_r, \mathbf{\Theta})]\right] - \text{D}_{\text{KL}}[q(\mathbf{\Psi})||p(\mathbf{\Psi} | \mathbf{\Theta})] ,
\end{split}
\end{align}
where $\text{D}_{\text{KL}}$ denotes the KL divergence, $\tilde{\boldsymbol{\Psi}}_r \sim q(\boldsymbol{\Psi})$, the minibatch input space is denoted by $\mathcal{I}_M$ and $N_{\text{MC}}$ is the number of Monte Carlo samples used to estimate $E_{q(\boldsymbol{\Psi})}$. $q(\boldsymbol{\Psi})$ and $p(\boldsymbol{\Psi} | \boldsymbol{\Theta})$ denote the approximate variational distribution and the prior distribution over the parameters respectively, both of which are assumed to be Gaussian in nature. A full derivation of this bound and the expression for the KL divergence between two normal distributions are included in the supplemental material.

We mirror the approach of \citet{cutajar2017random} and \citet{KingmaW13} by reparameterizing the weights and spectral frequencies, which allows for stochastic optimisation of the means and variances of our distributions over $\mathbf{W}$ and $\boldsymbol{\Omega}$ via gradient descent techniques. Specifically, we use the AdamW optimiser \citep{loshchilov2018decoupled}, implemented in PyTorch, as empirically it led to superior performance compared to other alternatives such as conventional stochastic gradient descent.

\subsection{Deep LFMs with variational inducing points} \label{sec:dlfm_pathwise}
Whilst the DLFM-RFF is computationally efficient and performs well empirically, Fourier feature-based GP approximations are prone to a phenomenon known as \textit{variance starvation}. As the number of observations in our training data increases, the predictions from the GP outside the domain of the training data become increasingly erratic, as the Fourier basis can only reliably represent stationary GPs, and the posterior is typically nonstationary \citep{wilson2020efficiently}. Motivated by this, we introduce a new model formulation and inference scheme for the DLFM, which we refer to as \textit{DLFM-VIP}, based on the variational inducing points framework for GP inference. This approach leverages the pathwise sampling method of \citet{wilson2020efficiently}, introduced in Section \ref{sec:background_pathwise}, which allows us to perform approximate Bayesian inference in our model in a doubly stochastic variational fashion. 

Firstly, we must reformulate our model in order to incorporate inducing points. We construct each layer of the DLFM-VIP using the following generative model,
\begin{equation}\label{eq:dlfm_vip_generative}
     \begin{split}
u_q(\mathbf{x}) &\sim \mathcal{GP}[0, k_{q}(\mathbf{x}, \mathbf{x}')], \qquad
         G^{(p)}_{d}(x_p) = e^{-\gamma_{d, p} x_p},\\
         G_{d}(\mathbf{x}) &=\prod_{p=1}^P G^{(p)}_{d}(x_p) , \qquad  G_{d, q}(\mathbf{x}) = a_{d, q} G_{d}(\mathbf{x}) , \\ 
&\qquad \mathbf{f}(\mathbf{x}) = \int_{-\infty}^{\mathbf{x}} \mathbf{G}(\mathbf{x}-\mathbf{z})\mathbf{u}(\mathbf{z})d\mathbf{z},
       \end{split}
\end{equation}
where $q=1,\dots, Q$, $p=1,\dots, P$ and $d=1,\dots, D$. By collecting all of the Green's functions corresponding to each distinct output and latent dimension, we form the matrix $\mathbf{G} \in \mathbb{R}^{D \times Q}$, and similarly we collect all of the latent functions to form the vector $\mathbf{u} \in \mathbb{R}^Q$. This model is similar to the NP-CGP proposed by \citet{mcdonald2022shallow}, however instead of using a fixed functional form for the filter, they place a GP over each filter $G_d^{(p)}$ in order to perform nonparametric kernel learning.

As discussed in Section \ref{sec:dlfm_rff_initial}, the initial conditions associated with our Green's function can constrain the layers of the DLFM-RFF in regions of input space close to the origin. Typically in LFMs, the convolution integral used to generate our outputs is computed with the limits $[0, \mathbf{x}]$, however by altering these limits to $[-\infty, \mathbf{x}]$ (as shown in Eq. \eqref{eq:dlfm_vip_generative}), we are able to circumvent the aforementioned initial condition problem entirely. In this formulation, samples from the model close to the origin may take any real value, and do not collapse to an initial condition at this point, therefore we maintain a rigorous quantification of uncertainty in this region of the input space.

\subsubsection{Output process sampling}
In order to compute samples from our output process, firstly we must produce functional samples from our latent processes $\{u\}_{q=1}^Q$. To achieve this, we use the pathwise sampling approach introduced by \citet{wilson2020efficiently}, which allows us to express functional samples from $u_q$ as follows,
\begin{align} \label{eq:latent_samples}
\begin{split}
    u_q^{(s)}(\cdot) = &\sum^{B}_{i=1}w_i\phi_i(\cdot) + k_{u_q, u_q}(\cdot, \mathbf{z}_q) \mathbf{K}_{u_q, u_q}^{-1}\left(\mathbf{v}_q -\boldsymbol{\Phi}\boldsymbol{w}\right).
\end{split}
\end{align}
In Eq. \ref{eq:latent_samples}, $\boldsymbol{\Phi} \in \mathbb{R}^{M \times B}$ is a matrix of features obtained by computing each of the $i=1, \dots, B$ basis functions at each of the $M$ inducing inputs. The individual basis functions are expressed by $\phi_i(\mathbf{x}) = \sqrt{2/B} \cos(\boldsymbol{\theta}_i^\top \mathbf{x} + \beta_i)$, where $\boldsymbol{\theta}_i \sim FT(k_{u_q, u_q})$ and $\beta_i \sim U(0, 2\pi)$, with $FT$ denoting the Fourier transform of the covariance and $U$ the uniform distribution. $\mathbf{w} \in \mathbb{R}^B$ is a vector of weights randomly sampled according to $w_i\sim\mathcal{N}(0, 1)$. $k_{u_{q}, u_{q}}$ refers to the EQ kernel of the latent process, and $\mathbf{z}_q \in \mathbb{R}^{M \times P}$ represents the $M$ inducing inputs, which have corresponding outputs $\mathbf{v}_q \in \mathbb{R}^{M \times 1}$. Finally, $\mathbf{K}_{u_{q}, u_{q}}$ denotes the covariance matrix obtained by evaluating the kernel function of the latent process for all combinations of our inducing inputs.

Now that we are able to compute functional samples for our latent processes, we can analytically map these through our convolution integral in order to generate samples from our output process,
\begin{align}
    \left(\mathbf{f}^{(s)} \mid \mathbf{V}^{\mathbf{u}} \right)(\mathbf{x}) &= \int_{-\infty}^{\mathbf{x}} \mathbf{G}(\mathbf{x}-\mathbf{z})\mathbf{u}^{(s)}(\mathbf{z})d\mathbf{z}.
\end{align}
Whilst omitted here for brevity, further details regarding the closed form result of this integration and its derivation are provided in the supplemental material. These multi-output samples can then be passed as input to the next layer of the model, allowing us to form a DLFM of arbitrary depth.

\subsubsection{Variational inference}
In order to perform approximate Bayesian inference in the DLFM-VIP, we once again employ stochastic variational inference. As we are now working in the inducing points framework, we follow the approach of \citet{salimbeni2017doubly}; this differs from the DLFM-RFF inference scheme outlined in Section \ref{sec:dlfm_rff_vi}, but solely in how we construct our joint and variational distributions. Firstly, let us collect all of the inducing variables at the $\ell$-th layer of the model into $\mathbf{V}^{\ell} = \{\mathbf{v}_q^{\ell} \}_{q=1}^Q$. Again, consider some input data $\mathbf{X} \in \mathbb{R}^{N\times P}$ with corresponding outputs $\mathbf{y} \in \mathbb{R}^{N\times D}$. We can write the joint distribution of the DLFM-VIP as,
\begin{align}
p(\mathbf{Y}, \{\mathbf{u}^\ell, \mathbf{V}^{\ell} \}^L_{\ell = 1}) =\prod^{N}_{i=1}  p(\mathbf{y}_{i} \mid \mathbf{F}_i^L)
\prod^{L}_{\ell=1} p(\mathbf{u}^\ell \mid \mathbf{V}^{\ell}) p(\mathbf{V}^{\ell}) ,
\end{align}
where the likelihood is given by $p(\mathbf{y}_{i}|\mathbf{F}_i^L) = \mathcal{N}(\mathbf{y}_i ; \mathbf{F}_i^L, \sigma^2_y)$, and $\mathbf{F}_i^L$ is the output of the model obtained from the $i$-th input. The latent GPs are independent such that $p(\mathbf{u}^\ell \mid \mathbf{V}^{\ell}) = \prod^{Q}_{q=1} p( u^{\ell}_{ q} \mid \mathbf{v}^{\ell}_{q})$, where $p( u^{\ell}_{q} \mid \mathbf{v}^{\ell}_{q})$ is a GP posterior and $p(\mathbf{V}^{\ell})$ represents the prior over the inducing points.

We construct our variational posterior as follows,
\begin{equation} \label{eq:dlfm_vip_variational_posterior}
    q(\{\mathbf{u}^\ell, \mathbf{V}^{\ell} \}^L_{\ell = 1}) = \prod^L_{\ell = 1} p(\mathbf{u}^\ell | \mathbf{V}^{\ell})q(\mathbf{V}^{\ell}),
\end{equation}
where $q(\mathbf{V}^{\ell}) =\prod^{Q_{\ell}}_{ q=1} \mathcal{N}(\mathbf{v}^{\ell}_{q};\boldsymbol{\mu}^{\ell}_{q}, \boldsymbol{\Sigma}^{\ell}_{q})$ is a variational distribution specific to the $\ell$-th layer, whose mean and covariance matrix are variational parameters. For the sake of clarity, we have omitted the factorisation over the layer dimensionality in the expression for the posterior. This allows us to compute the following stochastic approximation to our variational lower bound using $S$ Monte Carlo samples,
\begin{align} \label{eq:dlfm_vip_elbo}
    \mathcal{L} = \sum^N_{i=1} \frac{1}{S} \sum^S_{s=1} \log p(\mathbf{y}_i | \mathbf{F}_i^{L^{(s)}}) - \sum^L_{\ell = 1} \left[ D_\text{KL}[q(\mathbf{V}^{\ell}) \lVert p(\mathbf{V}^{\ell})] \right],
\end{align}
where $\mathbf{F}_i^{L^{(s)}}$ is a sample from the model. A full derivation of this bound is available in the supplemental material.

\section{Related Work} \label{sec:related}
Whilst the intersection of physics and deep learning has been a fruitful avenue for research for a number of years \citep{willard2020integrating}, to date there has been limited work on incorporating physical structure into deep \textit{probabilistic} models. The work of \citet{lorenzi2018constraining} is one example, however the authors achieve this by constraining the dynamics of their DGP with random feature expansions, rather than specifying a physics-informed kernel as in our approach. Additionally, the model developed by \citet{lorenzi2018constraining} is primarily designed for tackling the problem of ODE parameter inference in situations where the form of the underlying ODEs driving the behaviour of a system are known. In contrast, the DLFM does not assume any knowledge of the differential equations governing the given dynamical system being studied, and thus we do not attempt to perform parameter inference; our aim is to construct a robust, physics-inspired model with extrapolation capabilities and the ability to quantify uncertainty in its predictions.

\textit{Physics-informed deep kernel learning} (PI-DKL) \citep{wang2022physics} is another proposed approach to integrating physical priors into a probabilistic model. Unlike the work of \citet{lorenzi2018constraining} and in \textit{physics-informed neural networks} \citep{raissi2019physics}, PI-DKL is applicable to scenarios where we do not have complete knowledge regarding the form of the physical system we wish to model, as is our DLFM. PI-DKL uses differential operators to guide the deep kernel learning process, but the authors do not utilise the process convolution framework and pathwise sampling as we do, and nor is their model a deep GP, as the deep component is a deterministic neural network.

The new formulation of the DLFM which we introduce in Section \ref{sec:dlfm_pathwise} is conceptually similar to \citet{ross2021learning} and \citet{mcdonald2023nonparametric}, as both of these works map GP samples through a convolution integral analytically using pathwise conditioning \citep{wilson2020efficiently}. The first key difference between those works and our approach is that both \citet{ross2021learning} and \citet{mcdonald2023nonparametric} convolve one GP with another in order to obtain a more expressive GP kernel, whereas we convolve our latent GP with a Green's function in order to encode physical structure within the model. Additionally, both of the aforementioned works are shallow, single-layer GPs, whereas we present a deep model which can be composed of an arbitrary number of layers.

One final point we wish to clarify is that although the DLFM is a deep probabilistic model composed of a series of convolved GPs, our model differs from the convolutional DGP construction outlined by \citet{dunlop2018deep}, $f^{(\ell + 1)}(t) = \int f^{(\ell)}(t-\tau)u^{(\ell + 1)}(\tau)d \tau$, which the authors argue results in trivial behaviour with increasing depth. Instead, our model takes the form, $f^{(\ell + 1)}(f^{(\ell)}) = \int_0^{f^{(\ell)}} G(f^{(\ell)} - \tau^{(\ell + 1)})u^{(\ell + 1)}(\tau^{(\ell + 1)}) d \tau^{(\ell + 1)}$, which is more akin to the compositional construction outlined by \citet{dunlop2018deep}, just with a kernel which happens to involve a convolution.

\section{Experiments} \label{sec:experiments}
In this section we present experimental results which compare the empirical performance of both formulations of the DLFM to one another, and to prior work. All experiments using the DLFM-RFF were performed using the improvements discussed in Section \ref{sec:dlfm_rff_initial} and \ref{sec:dlfm_rff_reparam}, thus these results differ from those previously presented in \citet{mcdonald2021compositional}. Further details regarding the experimental setups are available in the supplemental material.

\subsection{Toy experiment} \label{sec:toy}
\begin{figure} 
    \centering
  \subfloat[GP \label{fig:toy_gp}]{%
       \includegraphics[width=0.33\linewidth]{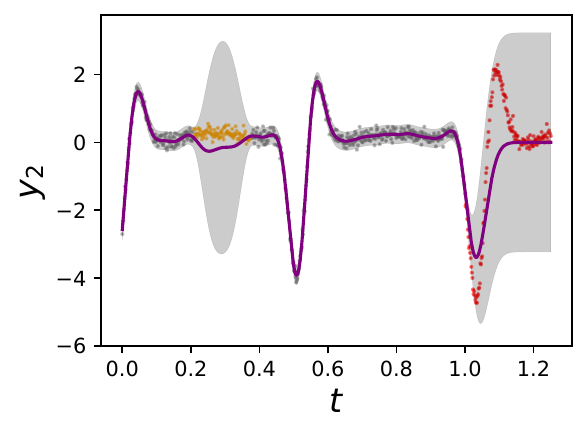}}
    \hfill
  \subfloat[DGP-DSVI \label{fig:toy_dgp_dsvi}]{%
        \includegraphics[width=0.33\linewidth]{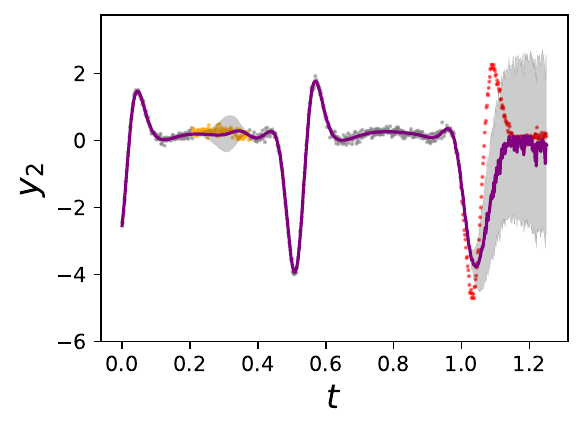}}
    \hfill
  \subfloat[DGP-RFF \label{fig:toy_dgp}]{%
        \includegraphics[width=0.33\linewidth]{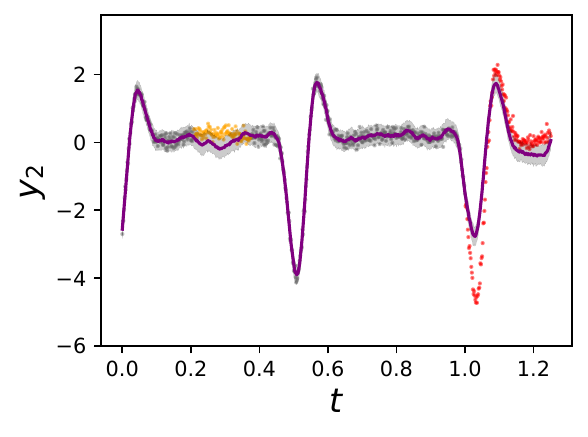}}
    \hfill
  \subfloat[DLFM-VIP \label{fig:toy_dlfm_vip}]{%
        \includegraphics[width=0.33\linewidth]{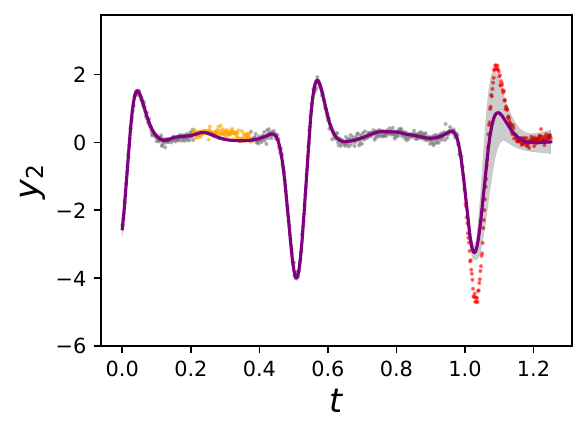}}
  \subfloat[DLFM-RFF \label{fig:toy_dlfm_rff}]{%
        \includegraphics[width=0.33\linewidth]{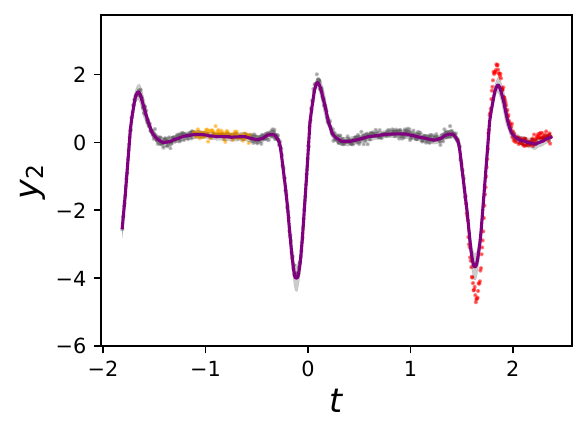}}
  \caption{Model fits to our toy dynamical system with latent function $u(t)=\cos(0.5 t) + 6\sin(3 t)$. Grey, orange and red data-points represent training, imputation test and extrapolation test data respectively. Purple curves represent predictive means, and shaded areas represent $\pm 2\sigma$.}
  \label{fig:toy_exp} 
\end{figure}
Firstly, to verify that our models are capable of modelling the compositional systems which their architectures resemble, 
we train the DLFM-RFF and DLFM-VIP on the inputs $t$ and noise-corrupted outputs $y_2(t)=f_2(t) + \epsilon$ of a system characterised 
by,
\begin{align}
\begin{split}
f_1(t) &= \int_0^t G_1(t-\tau)u(\tau)d\tau \\
f_2(f_1) &= \int_0^{f_1} G_2(f_1 - \tau^\prime) u(\tau^\prime) d\tau^\prime , \label{eq:toy_odes}
\end{split}
\end{align}
where $G_1(\cdot)$ and $G_2(\cdot)$ denote the Green's functions corresponding to the first order ODEs which these two integrals represent the solutions to. We evaluate an exact GP, a deep GP with random Fourier features (DGP-RFF) \citep{cutajar2017random}, a doubly stochastic variational deep GP (DGP-DSVI) \citep{salimbeni2017doubly}  and both the DLFM-RFF and DLFM-VIP. The latent function is sinusoidal in nature, resulting in a signal consisting of pulses with varying amplitude. From the predictions shown in Fig. \ref{fig:toy_exp}, we see that all five models capture the behaviour of the system well in the training regime, but only the deep models are able to extrapolate.

In both the imputation and extrapolation test regions, the DLFM-RFF visibly provides a better fit to the data than the DGP-RFF. By turning off reparameterization at test time, as discussed in Section \ref{sec:dlfm_rff_reparam}, the predictive distribution of the DLFM-RFF is not subject to the noise we see in the DGP-RFF predictions. The DLFM-VIP with 20 inducing points placed evenly across the entire domain is also capable of modelling the dynamics of the system with a more realistic quantification of uncertainty than the DGP-DSVI and DGP-RFF, however it does not capture the extremes of the function in the extrapolation regime as accurately as the DLFM-RFF is able to. 

The key takeaway from this experiment is that both the DLFM-RFF and DLFM-VIP exhibit superior short-range extrapolation ability compared to the DGP-RFF and DGP-DSVI respectively, evidencing the fact that our physics-informed kernel improves our ability to perform short-range extrapolation tasks such as this regardless of the inference scheme we select.
\begin{figure} 
  \centering
  \subfloat[DLFM-RFF \label{fig:imputation_rff}]{%
       \includegraphics[width=0.9\linewidth, trim={0 0 0 0.7cm}, clip]{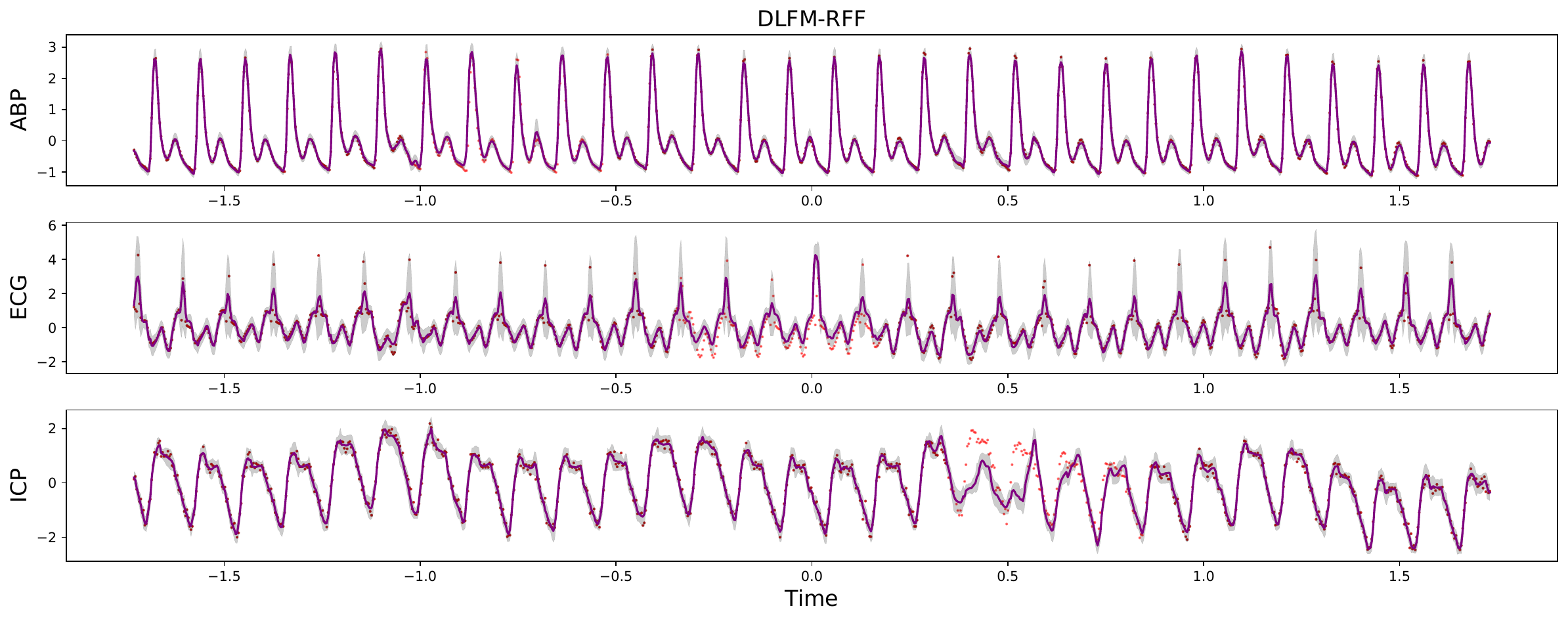}} \\
  \subfloat[DLFM-VIP \label{fig:imputation_vip}]{%
       \includegraphics[width=0.9\linewidth, trim={0 0 0 0.7cm}, clip]{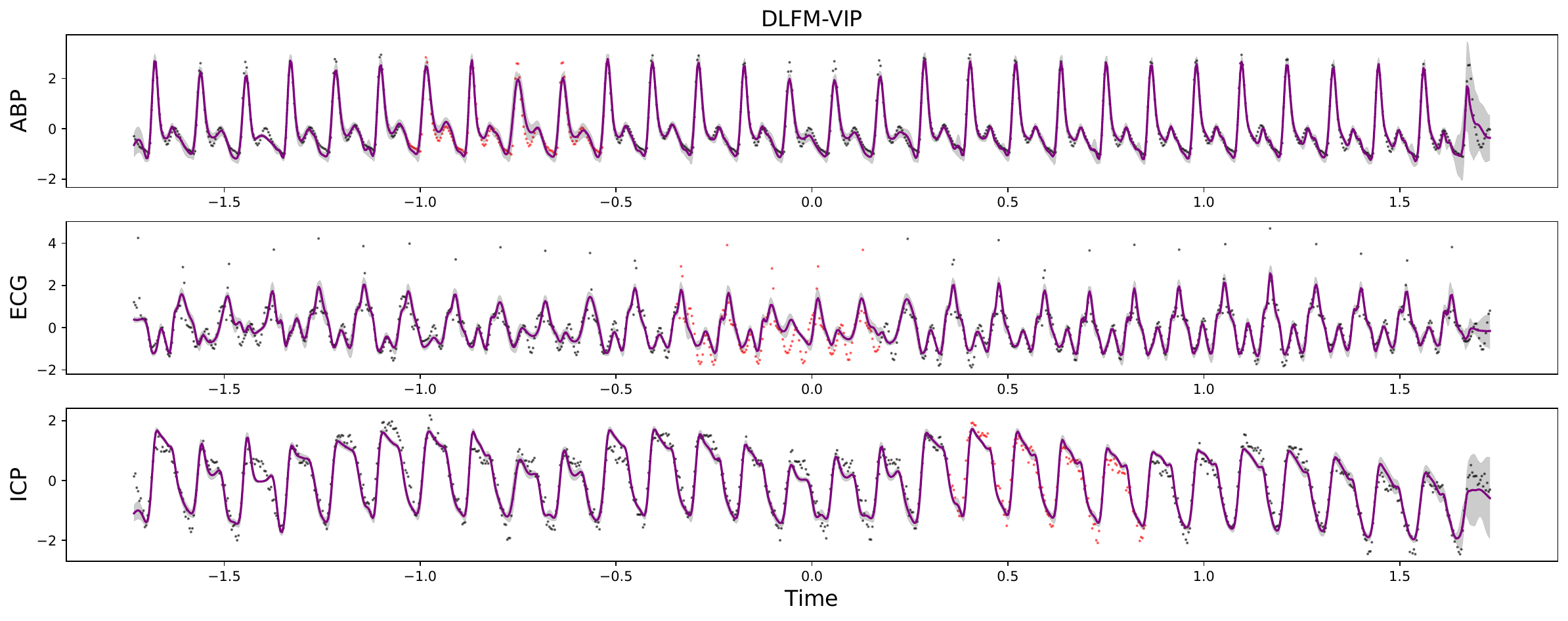}}
  \caption{Predictions generated for each of the three outputs within the CHARIS imputation experiment, from a DLFM-RFF and DLFM-VIP. Grey dots represent training data, orange dots represent test data, the purple line represents the predictive mean of the model, and the shaded grey areas in each plot represent $\pm 2\sigma$.}
  \label{fig:charis_imput} 
\end{figure}

\subsection{PhysioNet multivariate time series} \label{sec:experiments_physionet}
\begin{figure} 
  \centering
  \subfloat[DLFM-RFF \label{fig:extrapolation_rff}]{%
       \includegraphics[width=0.47\linewidth, trim={0 0 0 0.7cm}, clip]{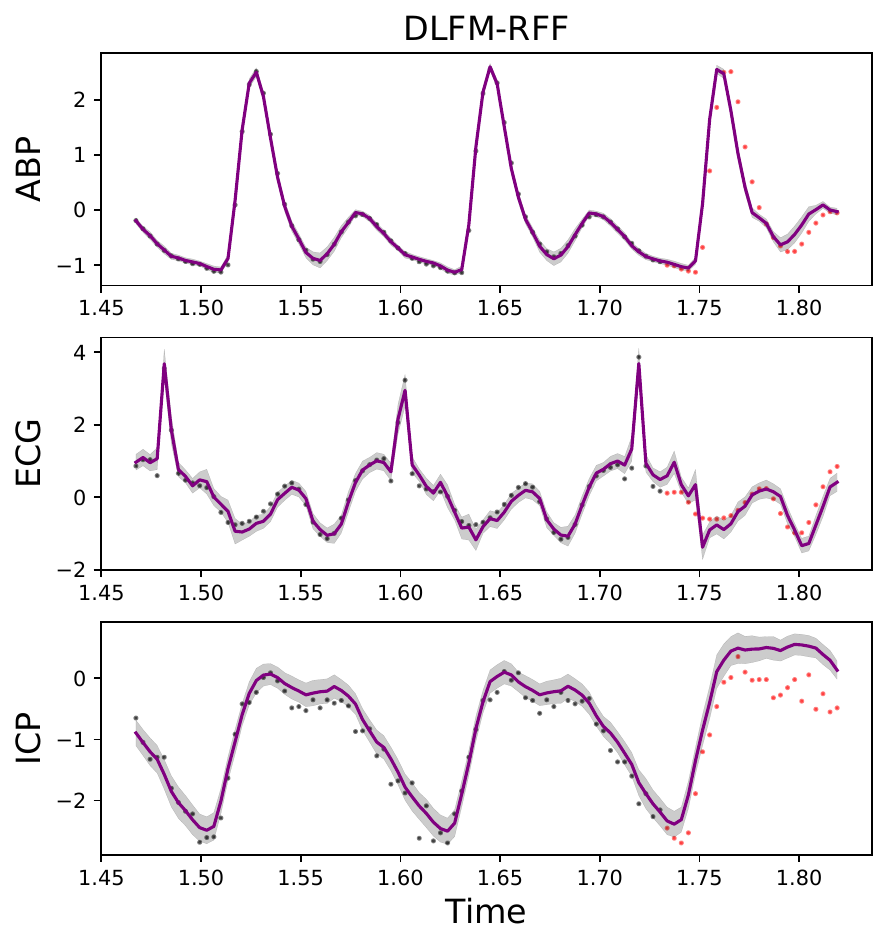}}
  \subfloat[DLFM-VIP \label{fig:extrapolation_vip}]{%
       \includegraphics[width=0.47\linewidth, trim={0 0 0 0.7cm}, clip]{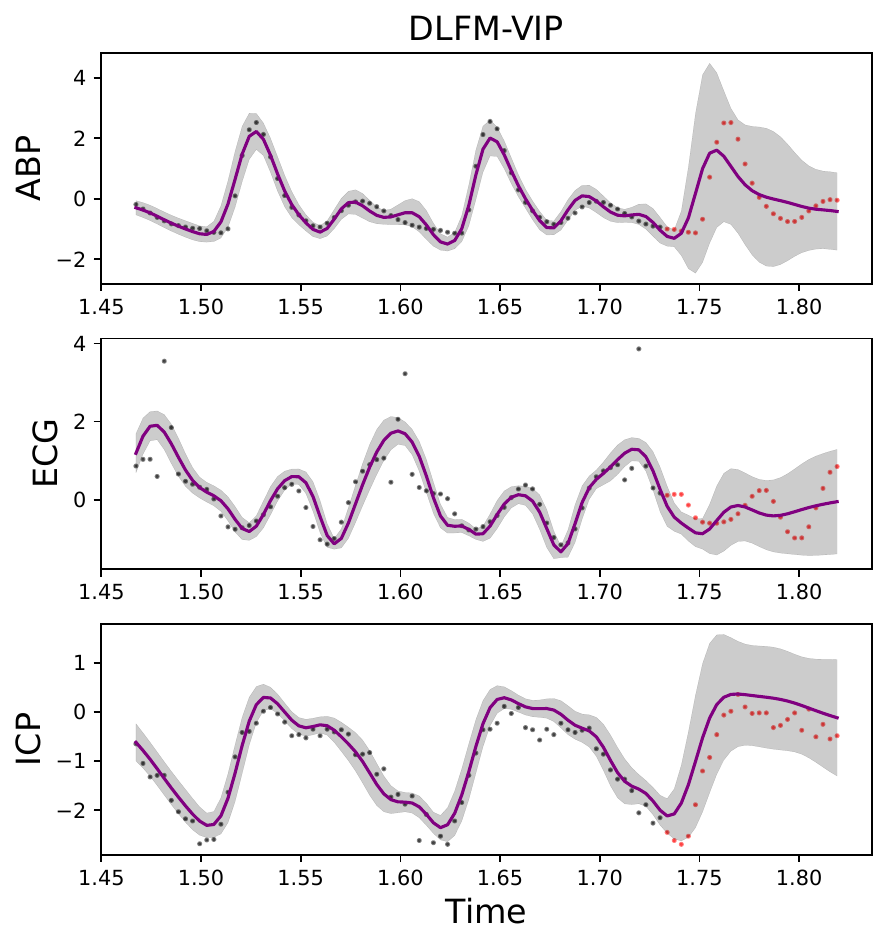}}
  \caption{Predictions generated for each of the three outputs within the CHARIS extrapolation experiment, from a DLFM-RFF and DLFM-VIP. Grey dots represent training data, orange dots represent test data, the purple line represents the predictive mean of the model, and the shaded grey areas in each plot represent $\pm 2\sigma$. Here, we have zoomed in on the final 100 timesteps which includes the extrapolation region; the previous 900 training observations are not shown.}
  \label{fig:charis_extrap} 
\end{figure}
To assess the ability of our model to capture highly nonlinear behaviour, we evaluate its performance on a subset of the CHARIS dataset (ODC-BY 1.0 License) \citep{kim2016trending}, which can be found on the PhysioNet data repository \citep{goldberger2000physiobank}. The data available for each patient consists of an electrocardiogram (ECG), alongside arterial blood pressure (ABP) and intracranial pressure (ICP) measurements; all three of these signals are sampled at regular intervals and are nonlinear in nature. We use a subset of this data consisting of the first 1000 time steps for a single patient. We focus on two predictive tasks. Firstly, we test the ability of our models to impute missing data by removing 150 non-overlapping datapoints from each output and using these as a test set. Additionally, we also consider the challenging task of extrapolating a short distance beyond the training input domain by training the aforementioned models on the first 975 observations and withholding the remaining 25 as a test set.
\begin{table}[!htb]
    \caption{Imputation test set results for each output within the CHARIS dataset, with standard error in brackets.}
    \centering
    \label{tab:charis_imput}
    \resizebox{\columnwidth}{!}{%
    \begin{tabular}{ccccccc}
        \toprule &
        \multicolumn{2}{c}{ABP} & \multicolumn{2}{c}{ECG} & \multicolumn{2}{c}{ICP} \\
        \cmidrule(r){2-3} \cmidrule(r){4-5} \cmidrule(r){6-7}
             & NMSE & MNLL & NMSE & MNLL & NMSE & MNLL \\
        \midrule
        DLFM-VIP & 0.10 (0.03) & 0.44 (0.08) & 0.34 (0.02) & 1.09 (0.04) & 0.16 (0.02) & 0.55 (0.06) \\
        DLFM-RFF & 0.05 (0.02) & 0.14 (0.32) & 0.33 (0.03) & 0.93 (0.04) & 0.09 (0.02) & 0.63 (0.24) \\
        DGP-DSVI & 0.38 (0.06) & 0.93 (0.07) & 0.38 (0.03) & 0.94 (0.04) & 0.31 (0.06) & 0.81 (0.08) \\
        DGP-RFF & 0.73 (0.02) & 1.18 (0.01) & 0.63 (0.02) & 1.21 (0.01) & 0.23 (0.02) & 0.73 (0.04) \\
        SLMC & 1.00 (0.001) & 1.42 ($\approx$ 0) & 1.00 ($\approx$ 0) & 1.43 ($\approx$ 0) & 1.03 (0.001) & 1.44 ($\approx$ 0) \\
        LFM-RFF & 1.00 (0.001) & - & 1.01 (0.01) & - & 0.97 (0.02) & - \\
        \bottomrule
      \end{tabular}
      }
\end{table}

\begin{table}[!htb]
    \caption{Extrapolation test set results for each output within the CHARIS dataset, with standard error in brackets.}
    \centering
    \label{tab:charis_extrap}
    \resizebox{\columnwidth}{!}{%
    \begin{tabular}{ccccccc}
        \toprule &
        \multicolumn{2}{c}{ABP} & \multicolumn{2}{c}{ECG} & \multicolumn{2}{c}{ICP} \\
        \cmidrule(r){2-3} \cmidrule(r){4-5} \cmidrule(r){6-7}
             & NMSE & MNLL & NMSE & MNLL & NMSE & MNLL \\
        \midrule
        DLFM-VIP & 0.65 (0.19) & 7.55 (1.96) & 1.30 (0.28) & 1.24 (0.17) & 0.45 (0.14) & 4.35 (1.19) \\
        DLFM-RFF & 0.11 (0.02) & 3.16 (2.14) & 0.83 (0.09) & 4.03 (1.84) & 0.39 (0.07) & 7.09 (3.65) \\
        DGP-DSVI & 1.00 ($\approx$ 0) & 1.82 ($\approx$ 0) & 1.45 (0.001) & 1.90 ($\approx$ 0) & 1.03 (0.03) & 1.49 (0.03) \\
        DGP-RFF & 1.49 (0.29) & 6.56 (1.32) & 4.54 (1.38) & 2.00 (0.47) & 0.47 (0.12) & 3.75 (0.83) \\
        SLMC & 1.04 (0.003) & 1.86 (0.009) & 1.52 (0.08) & 0.91 (0.02) & 1.05 (0.05) & 1.49 (0.04) \\
        \bottomrule
      \end{tabular}
      }
\end{table}
For the DLFM-RFF we use 100 random features, and for the DLFM-VIP we use 100 inducing points per layer. For both DLFMs, we use a single latent force per layer and the dimensionality of the hidden layers is set to be equal to the number of outputs in our dataset. As in the toy experiment, we once again compare our models to the DGP-RFF and DGP-DSVI. The parameters chosen for the DGP models align with the selections made for the DLFM as closely as possible in order to maintain parity. In addition to these deep models, we also evaluate the performance of a stochastic variational multi-output GP with 100 inducing points which uses the \textit{linear model of coregionalization} (SLMC) \citep{alvarez2012kernels}. Finally, we compare to a shallow LFM with random Fourier features derived from the same first order ODE as our model (LFM-RFF) \citep{guarnizo2018fast}, consisting of two latent forces and 20 random features; this setup was chosen as using more random features or latent forces did not appear to improve the performance of the LFM-RFF for this task. Given the poor performance of the LFM on the simpler imputation task, we do not include results for this model in the extrapolation setting.

Tables \ref{tab:charis_imput} and \ref{tab:charis_extrap} contain the results of the imputation and extrapolation experiments respectively. In order to evaluate and compare the performance of each technique, we compute the per-output test set MNLL for each model, as well as the normalised MSE (NMSE), which was chosen as it is a common metric for evaluating time series problems. We see from the results presented that the DLFM-RFF outperforms all of the baseline methods on the imputation task across all metrics, and yields a superior NMSE to all of the baselines on the extrapolation task. On the imputation task, the DLFM-VIP also outperforms all the baseline methods across both metrics, apart from the ECG MNLL for the DGP-DSVI. 

Overall the DLFM-VIP does not quite match the empirical performance of the DLFM-RFF on the imputation task, however the margin is relatively small and the DLFM-VIP does appear to be slightly more stable overall, not being as prone to the large standard errors which the DLFM-RFF can suffer from. On the extrapolation task however, the DLFM-RFF does significantly outperform the DLFM-VIP in terms of NMSE. Both the DLFM-RFF and DLFM-VIP achieve relatively poor test set MNLLs on the extrapolation task, often on par with or worse than the baseline methods. In the case of the DLFM-RFF this can likely be attributed to variance starvation; it is clear from Figure \ref{fig:charis_extrap} that the confidence intervals are poorly calibrated given that they do not cover a significant portion of the test data. Conversely, for the DLFM-VIP, these performance issues arise because whilst the confidence intervals are arguably more realistic, the model is less capable of extrapolating the dynamics outside of the test region, as is evident from comparing the predictive means of the two DLFMs. These extrapolation results present an interesting opportunity for future work, which we will discuss in further detail in Section \ref{sec:dlfm_discussion_inference}.

\subsection{UCI regression} \label{sec:uci}
As noted in \citet{mcdonald2021compositional}, whilst the DLFM was not originally designed with tabular regression problems in mind, the model is surprisingly effective in this setting, likely due to the factors discussed in Section \ref{sec:dlfm_discussion}. We consider two single-output UCI regression datasets commonly used for benchmarking deep probabilistic models, \textit{power} ($N=9568$, $D=4$) and \textit{protein} ($N=45730$, $P=9$). Again, we evaluate both of our models alongside the DGP-DSVI and DGP-RFF, however for the shallow GP benchmark we now use a stochastic variational GP (SVGP) with an EQ kernel and 100 inducing points.

From the results shown in Table \ref{tab:uci} we see that on the \textit{power} dataset, both variants of the DLFM outperform the shallow benchmark in terms of both metrics, and also provide comparable results to the DGP-RFF, with slightly improved MNLL. On the \textit{protein} dataset, the DLFM-RFF is unable to exceed the performance of the DGP-RFF or shallow benchmark, however the new DLFM-VIP improves upon this, exceeding the shallow benchmark in terms of both RMSE and MNLL. The DGP-DSVI is the best performing approach across both datasets and metrics. As mentioned above, the model was not devised for application to these types of problems, however the relatively strong performance of the DLFMs in this setting is related to some theoretical insights on the relationship between the DLFM and conventional DGPs, which we will discuss in Section \ref{sec:dlfm_discussion}.

\begin{table}
    \centering
    \caption{Test set results on the UCI benchmarks, with standard error in brackets.}
    \label{tab:uci}
      \begin{tabular}{ccccc}
        \toprule &
        \multicolumn{2}{c}{power} & \multicolumn{2}{c}{protein} \\
        \cmidrule(r){2-3} \cmidrule(r){4-5}
             & RMSE & MNLL & RMSE & MNLL \\
        \midrule
        DLFM-VIP & 3.84 (0.03) & 2.77 (0.01) & 4.21 (0.02) & 2.86 (0.005) \\
        DLFM-RFF & 3.83 (0.03) & 2.78 (0.01) & 4.67 (0.03) & 2.97 (0.007) \\
        DGP-DSVI & 3.66 (0.04) & 2.72 (0.01) & 3.96 (0.01) & 2.79 (0.003) \\
        DGP-RFF & 3.84 (0.05) & 2.80 (0.01) & 4.03 (0.01) & 2.82 (0.003) \\
        SVGP & 3.86 (0.04) & 2.77 (0.01) & 4.33 (0.01) & 2.89 (0.003) \\
        \bottomrule
      \end{tabular}
\end{table}

\section{Model analysis \& discussion } \label{sec:dlfm_discussion}
Both the DLFM-RFF and DLFM-VIP explicitly encode the dynamics of a linear first order ordinary differential equation within the kernel of each of their GP layers. Despite this enforced inductive bias, the results presented in Section \ref{sec:uci} show that the DLFM still retains the generality of a conventional DGP, and achieves comparable performance on tasks where we would not necessarily expect this inductive bias to be of any use. In this section, we briefly discuss how this behaviour can be explained.

Firstly, we can explain this behaviour by considering the convolutional nature of the DLFM. In a regular DGP, the layers of the model warp the inputs, and the DLFM performs the exact same process. However, in the DLFM, due to the convolution of the GP at each layer of the model with the Green's function, we introduce an additional degree of flexibility which allows us to control the nature of the warping which occurs at each layer. The Green's function can be interpreted as a filter over the GPs at each layer of the model, which not only allows us to warp the intermediate inputs, but also gives us the ability to selectively filter out certain frequencies of the latent GPs, altering their degree of smoothness. It is notoriously challenging to reason about a sensible choice for the kernel in the inner layers of a DGP \citep{mcdonald2022shallow}; this is further motivation for our usage of the ODE1 kernel, as it effectively just provides us with a more flexible approach to learning the dynamics of the latent space. However, we can also look to interpret these findings in a more explicit manner by exploring the connection between the ODE1 and EQ kernels. 

\subsection{Decay parameter analysis}
As discussed above, the ODE1 kernel adds useful structure to the model when required. However, this additional complexity on top of the EQ kernel of the latent GP may not always be required, thus an advantageous quality in our model would be the ability to adaptively revert to an EQ-like kernel when the physics-informed dynamics are not required. Consider the scenario where instead of using a Green's function in our convolution integral, we use a \textit{Dirac delta function}. With simplified notation for ease of exposition, this yields the result $f(x) = \int_{-\infty}^x \delta(x-\tau)u(\tau)d\tau = H(x)u(x)$, where $H(\cdot)$ is the Heaviside step function. As the delta function is analogous to an identity function when working with convolutions, we have simply recovered our original latent GP with EQ covariance. 
\begin{figure} 
  \centering
  \includegraphics[width=0.75\linewidth]{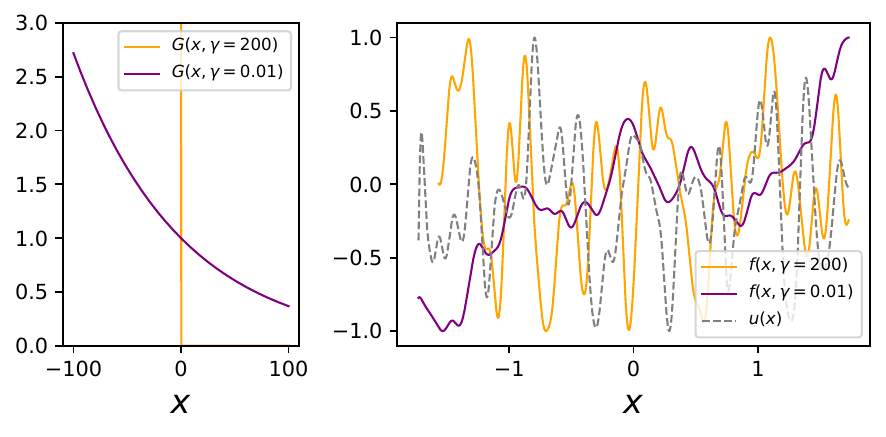}
  \caption{On the left we see the Green's function visualised for two different values of $\gamma$, with $G(x)$ resembling a delta function as $\gamma$ is increased. The effect of this can be seen on the right, where we have a sample from our latent GP, alongside output samples from a single layer of the DLFM-VIP for two different values of $\gamma$. For $\gamma=200$, the sample resembles the sample from the latent GP with EQ kernel, whereas for $\gamma=0.01$, the filter has a much more visible effect on the form of the sample.}
  \label{fig:gamma_dirac}
\end{figure}

As shown in Figure \ref{fig:gamma_dirac}, as $\gamma \to \infty$, our ODE1 Green's function $e^{-\gamma x}$ forms an increasingly close approximation to a delta function, subject to a normalisation constant. This means that our DLFM behaves increasingly like a conventional DGP with EQ kernels as $\gamma$ increases, and as $\gamma$ is a freely optimizable parameter, the model can effectively select the extent to which the physics-informed inductive bias of the ODE1 kernel is required. In Figure \ref{fig:gamma_dirac}, the prior samples we show have been normalised to the range $[-1, 1]$ for ease of comparison; the model would be able to perform this rescaling during training through optimisation of the amplitude parameters present in each LFM layer. As previously alluded to, one caveat to note is that whilst our Green's function approximates a delta function, it does not \textit{tend} to a delta function as it does not adhere to the normalisation to unity constraint which a delta function must satisfy, as $\int_{-\infty}^{\infty} H(t) e^{-\gamma x} dx = \int_{0}^{\infty} e^{-\gamma x} dx = 1/\gamma$.

\subsection{Inference scheme comparison} \label{sec:dlfm_discussion_inference}
We present experimental results in this work which involve making predictions in three very distinct scenarios. Firstly, the UCI experiments are an \textit{interpolation} problem, where the test data used to evaluate each model is randomly sampled from across the input domain. From the results shown in Table \ref{tab:uci}, we can see that the DLFM-VIP fares well in this setting, outperforming the DLFM-RFF. Secondly, we have considered an \textit{imputation} problem with the PhysioNet data, where the test data consists of non-overlapping chunks of contiguous data-points from each individual output signal. This is a challenge as the model has not been exposed to any data from these regions during training, however the model has had access to some contextual information in the form of the data-points either side of the test region, and the values of the other outputs within the test region. From the results shown in Table \ref{tab:charis_imput}, we see that the DLFM-VIP exhibits almost comparable performance to the DLFM-RFF in this setting, and outperforms the baselines tested. Finally, we considered an \textit{extrapolation} problem with the PhysioNet data, where the test data is a region entirely outside of the training domain, and the model has not has access to any of the output values within this region. As we can see from the results in Table \ref{tab:charis_extrap} and Figure \ref{fig:charis_extrap}, it is in this setting where the DLFM-VIP begins to struggle, failing to outperform the DLFM-RFF. This occurs regardless of whether or not we choose to fix evenly spaced inducing inputs across the entirety of the training \textit{and} test domain at the first layer of the DLFM-VIP.

As noted by \citet{rudner2020inter}, inducing points-based GP inference methods create \textit{local} approximations to the target function, which considerably limits the ability of such models to capture, and thus extrapolate, global structure within data. The two primary aims of this work were firstly to improve the interpolation abilities of the DLFM by addressing the issues associated with RFF-based posteriors, and secondly to assess whether the LFM framework would be able to overcome the aforementioned difficulties associated with extrapolating with inducing points-based models. The first aim was achieved, however we find empirically that the DLFM-VIP is only capable of performing pure extrapolation for short distances outside of the input domain, and does so with a higher degree of error compared to the DLFM-RFF. 

The relative empirical effectiveness of the DLFM-RFF in the extrapolation regime may seem at odds with the fact that, as discussed in Section \ref{sec:dlfm_pathwise}, variance starvation in Fourier feature-based GPs can cause problems outside of the training domain when the number of data-points is larger than the number of Fourier features used (which is generally the case in our experiments). We posit that the global structure imposed by the physics-informed filter $G(\cdot)$ exhibits a degree of control over the usually erratic mean predictions which can result when attempting to extrapolate with Fourier feature-based GPs experiencing variance starvation. However, the filters do not have any means of counteracting the over-confidence \citep{calandriello2019gaussian} which is characteristic of variance starvation, which explains the low variance in the extrapolation predictions returned by the DLFM-RFF, and the poor test set log-likelihoods on the PhysioNet extrapolation task discussed in Section \ref{sec:experiments_physionet}.

In summary, whilst the DLFM-RFF and DLFM-VIP have different strengths and weaknesses, they both represent a notable improvement over their non-physics-informed DGP counterparts. A pressing issue for future work is how we can combine the advantages of these two models to obtain a model which is capable of performing accurate long-range extrapolation whilst avoiding the issue of variance starvation. In Section \ref{sec:conclusion}, we briefly discuss a possible direction for future work based on interdomain DGPs which may alleviate this issue entirely.

\section{Conclusion \& Future Work} \label{sec:conclusion}
In this work we outline the \textit{deep latent force model} (DLFM), a generalised framework for performing Bayesian deep learning with physics-informed Gaussian processes. Our results show that DLFMs are able to effectively capture nonlinear dynamics in both single and multiple-output settings, as well as achieve competitive performance on benchmark tabular regression tasks. Two formulations of the DLFM are presented based on random Fourier features and variational inducing points, which are more suited to extrapolation and interpolation tasks respectively, for reasons discussed in Section \ref{sec:dlfm_discussion}.

We envision a number of avenues for future research. Firstly, the inference procedure in both models could be modified to extend the variational treatment we consider to encompass the kernel hyperparameters. Additionally, as discussed earlier, improving the extrapolation abilities of the DLFM-VIP is a key problem. Prior work on shallow LFMs with inducing points-based approximations has used periodic latent processes to address the latter problem \citep{moss2021approximate}, however this is not an option in the pathwise sampling setting, as the resulting convolution integral has no closed form solution. Another kernel-based solution may be to instead derive a kernel from a 2nd order ODE in order to imbue the model with more periodic long-range structure. This approach would however be more computationally intensive, and introduce more hyperparameters to estimate, which may necessitate a less biased, fully Bayesian approach to estimating these hyperparameters.

A non-trivial but potentially highly fruitful solution to this issue could involve fundamentally altering the structure of the model to mirror that of \citet{rudner2020inter}, who use an interdomain approach with RKHS Fourier features in order to improve the ability of DGPs to capture global structure in data. This would in effect combine the advantages of the DLFM-RFF and the DLFM-VIP, resulting in a model which can extrapolate dynamics globally whilst avoiding the problems associated with using a purely Fourier feature-based posterior.




\newpage

\appendix
\section{Derivation of DLFM-VIP Output Samples}
To draw samples from the DLFM-VIP, we initially sample from the variational distribution over our inducing points, $\mathbf{V}^{\mathbf{u}^{(\ell)}} \sim q(\mathbf{V}^{\mathbf{u}^{(\ell)}})$, before using the pathwise sampling method of \citet{wilson2020efficiently} to sample latent functions $\mathbf{u}^{(s)}$. We then map these functions analytically through the convolution integral, splitting the multi-dimensional integrals into products of one-dimensional integrals to allow for multi-dimensional inputs, in a similar fashion to the approach taken by \citet{mcdonald2022shallow}. If we consider a kernel derived from a first order ODE, such that $G_d^{(p)}(x_p) = e^{-\gamma_{d, p}x_p}$, we can compute the closed form expression for a sample function from a given layer as follows,
{\allowdisplaybreaks
\begin{align*}
\Bigl(&\mathbf{f}^{(s)} \mid \mathbf{V}^{\mathbf{u}} \Bigr)(\mathbf{x}) = \int_{\mathbf{-\infty}}^{\mathbf{x}} \mathbf{G}(\mathbf{x} - \boldsymbol{\tau}) \mathbf{u}^{(s)}(\boldsymbol{\tau}) d\boldsymbol{\tau}
\\ &= \int_{\mathbf{-\infty}}^{\mathbf{x}} a e^{-\boldsymbol{\gamma^\top}(\mathbf{x}-\boldsymbol{\tau})} \left(\boldsymbol{\Phi}(\boldsymbol{\tau}) \mathbf{w} + k\left(\boldsymbol{\tau}, \mathbf{z}\right) \mathbf{q} \right) d \boldsymbol{\tau}
\\ &= a \int_{\mathbf{-\infty}}^{\mathbf{x}} e^{-\boldsymbol{\gamma^\top}(\mathbf{x}-\boldsymbol{\tau})} \left(\sum^{B}_{i=1} w_i \phi_i (\boldsymbol{\tau}) \ + \ \sum_{j=1}^M q_j k(\boldsymbol{\tau}, \mathbf{z}_{j}) \right) d\boldsymbol{\tau}
\\ &= a \sqrt{\frac{2}{B}} \sum^{B}_{i=1} w_i \int_{\mathbf{-\infty}}^{\mathbf{x}} e^{-\boldsymbol{\gamma^\top}(\mathbf{x}-\boldsymbol{\tau})} \cos\left(\boldsymbol{\theta}_i^\top \boldsymbol{\tau} + \beta_i \right) d\boldsymbol{\tau}  + \sum_{j=1}^M q_j \int_{\mathbf{-\infty}}^{\mathbf{x}} e^{-\boldsymbol{\gamma^\top}(\mathbf{x}-\boldsymbol{\tau})} k(\boldsymbol{\tau}, \mathbf{z}_{j}) d\boldsymbol{\tau}
\\ &= a \sqrt{\frac{2}{B}} \sum^{B}_{i=1} w_i \int_{\mathbf{-\infty}}^{\mathbf{x}} e^{-\boldsymbol{\gamma^\top}(\mathbf{x}-\boldsymbol{\tau})} \left( \frac{e^{j \left(\boldsymbol{\theta}_i^\top \boldsymbol{\tau} + \beta_i \right)} + e^{-j \left(\boldsymbol{\theta}_i^\top \boldsymbol{\tau} + \beta_i \right)}}{2} \right) d\boldsymbol{\tau} \  \\ & \qquad \quad + \sum_{j=1}^M q_j \int_{\mathbf{-\infty}}^{\mathbf{x}} e^{-\boldsymbol{\gamma^\top}(\mathbf{x}-\boldsymbol{\tau})} k(\boldsymbol{\tau}, \mathbf{z}_{j}) d\boldsymbol{\tau}
 \\ &= a  \sqrt{\frac{2}{B}} \sum^{B}_{i=1} w_i \int_{\mathbf{-\infty}}^{\mathbf{x}} \left( \prod_{p=1}^P e^{-\gamma_p (x_p-\tau_p)} \right) \left( \frac{e^{j\beta_i} e^{\sum_{p=1}^P j \theta_{i,p} \tau_p } + e^{-j\beta_i} e^{\sum_{p=1}^P -j \theta_{i,p} \tau_p } }{2} \right) d\boldsymbol{\tau} \\
 & \qquad + a \sum_{j=1}^M q_j \int_{\mathbf{-\infty}}^{\mathbf{x}} \left( \prod_{p=1}^P e^{-\gamma_p (x_p-\tau_p)} \right) \ \left( \prod_{p=1}^P \sigma^2 e^{-\frac{\left(\tau_p - {z_{j, p}}^\prime \right)^2}{2\ell_p^2}} \right) d\boldsymbol{\tau}
\\ &= a  \sqrt{\frac{2}{B}} \sum^{B}_{i=1} w_i \int_{\mathbf{-\infty}}^{\mathbf{x}} \left( \prod_{p=1}^P e^{-\gamma_p (x_p-\tau_p)} \right) \left( \frac{e^{j\beta_i} \prod_{p=1}^P e^{j \theta_{i,p} \tau_p } + e^{-j\beta_i} \prod_{p=1}^P e^{-j \theta_{i,p} \tau_p } }{2} \right) d\boldsymbol{\tau} \\
 & \qquad + a \sum_{j=1}^M q_j \int_{\mathbf{-\infty}}^{\mathbf{x}} \prod_{p=1}^P e^{-\gamma_p (x_p-\tau_p)} \ \sigma^2 e^{-\frac{\left(\tau_p - {z_{j, p}}^\prime \right)^2}{2\ell_p^2}} d\boldsymbol{\tau}
\\ &= a  \sqrt{\frac{2}{B}} \sum^{B}_{i=1} w_i \Biggl( \frac{e^{j\beta_i}}{2} \prod_{p=1}^P  \int_{-\infty}^{x_p}  e^{j \theta_{i,p} \tau_p -\gamma_p (x_p-\tau_p)} d\tau_p + \frac{e^{-j\beta_i}}{2} \prod_{p=1}^P  \int_{-\infty}^{x_p}  e^{-j \theta_{i,p} \tau_p -\gamma_p (x_p-\tau_p)} d\tau_p \Biggr) \\
 & \qquad + a\sigma^2 \sum_{j=1}^M q_j \prod_{p=1}^P \int_{-\infty}^{x_p} e^{-\gamma_p (x_p-\tau_p)} \ e^{-\frac{\left(\tau_p - {z_{j, p}}^\prime \right)^2}{2\ell_p^2}} d\tau_p
\\ &= a  \sqrt{\frac{2}{B}} \sum^{B}_{i=1} w_i \left( \frac{e^{j\beta_i}}{2} \prod_{p=1}^P  \frac{e^{j\theta_{i,p}x_p}}{\gamma_p + j\theta_{i,p}} + \frac{e^{-j\beta_i}}{2} \prod_{p=1}^P  \frac{e^{-j\theta_{i,p}x_p}}{\gamma_p - j\theta_{i,p}} \right) \\
 & \qquad + a\sigma^2 \sum_{j=1}^M q_j \prod_{p=1}^P  \ell_p \sqrt{\frac{\pi}{2}} e^{\frac{\gamma_p}{2} \left(\gamma_p \ell_p^2 + 2z_{p, j} - 2x_p \right)} \text{erfc}\left[\frac{\gamma_p \ell_p^2 + z_{p, j} - x_p}{\ell_p \sqrt{2}} \right]
\end{align*}
}
where for ease of exposition, from the second line onwards we consider the simplest case, where $D=1$ and $Q=1$; the expressions do trivially generalise to scenarios where both are greater than one (as does our implementation). $q_j$ represents the $j$-th element of the vector $\mathbf{q}=\mathbf{K}_{u, u}^{-1} \left(\mathbf{v}^{u} - \boldsymbol{\Phi} \mathbf{w}\right)$, which is constant with respect to $\boldsymbol{\tau}$ in the above integral. The vector $\mathbf{w} \in \mathbb{R}^{B}$ in this expression consists of entries $w_i \sim \mathcal{N}(0, 1)$. Our $i=1, \dots, B$ \textit{original} basis functions are denoted by $\phi_i(\boldsymbol{\tau}) = \sqrt{2/B}\cos(\boldsymbol{\theta}_i^\top \boldsymbol{\tau} + \beta_i)$, where $\beta_i \sim U(0, 2\pi)$ and $\boldsymbol{\theta_i} \sim FT(k)$, with $FT(k)$ denoting the Fourier transform of the covariance of the latent process. These basis functions can be collected into a matrix denoted by $\boldsymbol{\Phi}(\boldsymbol{\tau}) \in \mathbb{R}^{M \times B}$. \textit{erfc} is used to denote the complementary error function.

A point of note regarding our implementation of the DLFM-VIP is that we choose to share the inducing inputs and kernel hyperparameters across the $Q$ latent processes. This reduces the computational complexity, and therefore runtime, of the model, however this restriction could be relaxed in scenarios where more flexibility is desired.

\subsection{Verifying the real nature of the samples}
The final expression shown above contains complex quantities within the following term,
\begin{align*}
\frac{e^{j\beta_i}}{2} \prod_{p=1}^P  \frac{e^{j\theta_{i,p}x_p}}{\gamma_p + j\theta_{i,p}} + \frac{e^{-j\beta_i}}{2} \prod_{p=1}^P  \frac{ e^{-j\theta_{i,p}x_p}}{\gamma_p - j\theta_{i,p}} .
\end{align*}
We can simplify this term, and in the process verify that it is in fact real-valued, using the properties of conjugate complex numbers. The expression can also be written as,
\begin{align*}
\frac{1}{2} \prod_{p=1}^P  \underbrace{\frac{e^{j(\theta_{i,p}x_p+\beta_i)}}{\gamma_p + j\theta_{i,p}}}_A+ \frac{1}{2} \prod_{p=1}^P  \underbrace{\frac{e^{-j(\theta_{i,p}x_p +\beta_i)}}{\gamma_p - j\theta_{i,p}}}_B
\end{align*}
The numerator in term B is the complex conjugate of the numerator in term A. Likewise, the denominator in term B is the complex conjugate of the denominator in term A. We can then write,
\begin{align*}
\frac{1}{2} \prod_{p=1}^P  \frac{e^{j(\theta_{i,p}x_p+\beta_i)}}{\gamma_p + j\theta_{i,p}}+ \frac{1}{2} \prod_{p=1}^P  \frac{\left(e^{j(\theta_{i,p}x_p +\beta_i)}\right)^*}{\left(\gamma_p + j\theta_{i,p}\right)^*} .
\end{align*}
Using the properties $z_1^*/z_2^* =(z_1/z_2)^*$ and $z_1^*z_2^*=(z_1z_2)^*$, the expression above simplifies to
\begin{align*}
\frac{1}{2} \left\{\prod_{p=1}^P  \frac{e^{j(\theta_{i,p}x_p+\beta_i)}}{\gamma_p + j\theta_{i,p}}+ \left[\prod_{p=1}^P  \frac{e^{j(\theta_{i,p}x_p +\beta_i)}}{\gamma_p + j\theta_{i,p}}\right]^*\right\}.
\end{align*}
Finally, using $z_1+z_1^*=2\text{Re}(z_1)$, we can write,
\begin{align*}
\frac{e^{j\beta_i}}{2} \prod_{p=1}^P  \frac{e^{j\theta_{i,p}x_p}}{\gamma_p + j\theta_{i,p}} + \frac{e^{-j\beta_i}}{2} \prod_{p=1}^P  \frac{e^{-j\theta_{i,p}x_p}}{\gamma_p - j\theta_{i,p}}=\text{Re}\left[\prod_{p=1}^P  \frac{e^{j(\theta_{i,p}x_p+\beta_i)}}{\gamma_p + j\theta_{i,p}}\right].
\end{align*}
This in turn allows us to simplify the final expression for our output samples to the following,
\begin{align*}
    \begin{split}
        \Bigl(\mathbf{f}^{(s)} &\mid \mathbf{V}^{\mathbf{u}} \Bigr)(\mathbf{x}) = a  \sqrt{\frac{2}{B}} \sum^{B}_{i=1} w_i \text{Re}\left[\prod_{p=1}^P  \frac{e^{j(\theta_{i,p}x_p+\beta_i)}}{\gamma_p + j\theta_{i,p}}\right] \\ & \qquad \qquad \qquad + a\sigma^2 \sum_{j=1}^M q_j \prod_{p=1}^P  \ell_p \sqrt{\frac{\pi}{2}} e^{\frac{\gamma_p}{2} \left(\gamma_p \ell_p^2 + 2z_{p, j} - 2x_p \right)} \text{erfc}\left[\frac{\gamma_p \ell_p^2 + z_{p, j} - x_p}{\ell_p \sqrt{2}} \right].
    \end{split}
\end{align*}

\section{Variational lower bound derivations}
In this section we present derivations of the variational lower bounds used for performing stochastic variational inference in both the DLFM-RFF and the DLFM-VIP. The Kullback-Leibler (KL) divergence between two normal distributions $p_A\left(x \;\middle|\; \mu_A, \sigma_A^2 \right) = \mathcal{N}\left(x \;\middle|\; \mu_A, \sigma_A^2 \right)$ and $p_B\left(x \;\middle|\; \mu_B, \sigma_B^2 \right) = \mathcal{N}\left(x \;\middle|\; \mu_B, \sigma_B^2 \right)$, appears in both of these derivations, and can be expressed in closed form as,
\begin{align*}
\text{D}_\text{KL}\left[p_A(x) \mid \mid p_B(x) \right] = \frac{1}{2} \left[\log \left(\frac{\sigma_B^2}{\sigma_A^2} \right) -1 + \frac{\sigma_A^2}{\sigma_B^2} + \frac{(\mu_A - \mu_B)^2}{\sigma_B^2} \right] .
\end{align*}

\subsection{DLFM-RFF}
Denoting $\bm{\Psi} = \left\{\bm{W}, \bm{\Omega} \right\}$ for ease of notation, we assume a factorised prior over the spectral frequencies and weights across all layers, which takes the form,
\begin{align*}
p(\bm{\Psi}|\bm{\Theta}) = \prod_{\ell=0}^{L-1} p\left(\bm{\Omega}^{(\ell)} \mid \bm{\Theta}^{(\ell)}\right)  p\left(\bm{W}^{(\ell)}\right) = \prod_{ij\ell} q\left(\Omega_{ij}^{(\ell)}\right) \prod_{ij\ell} q\left(W_{ij}^{(\ell)}\right) ,
\end{align*}
where, 
\begin{align*}
q\left(\Omega_{ij}^{(\ell)} \;\middle|\; m_{ij}^{(\ell)}, (s^2)_{ij}^{(\ell)} \right) &= \mathcal{N}\left(\Omega_{ij}^{(\ell)} \;\middle|\; m_{ij}^{(\ell)}, (s^2)_{ij}^{(\ell)}\right) \\
q\left(W_{ij}^{(\ell)} \;\middle|\; \mu_{ij}^{(\ell)}, (\beta^2)_{ij}^{(\ell)} \right) &= \mathcal{N}\left(W_{ij}^{(\ell)} \;\middle|\; \mu_{ij}^{(\ell)}, (\beta^2)_{ij}^{(\ell)}\right).
\end{align*} 
As discussed in the main paper, we reparameterise the spectral frequencies and weights as,
\begin{align*}
\left(\tilde{\Omega}_r^{(\ell)}\right)_{ij} = s_{ij}^{(\ell)} \epsilon_{rij}^{(\ell)} + m_{ij}^{(\ell)} \\
\left(\tilde{W}_r^{(\ell)}\right)_{ij} = \beta_{ij}^{(\ell)} \epsilon_{rij}^{(\ell)} + \mu_{ij}^{(\ell)} 
\end{align*}
where $\epsilon_{rij}^{(\ell)}$ represent samples from a standard normal, $\tilde{\Omega}_r \sim q\left(\Omega_{ij}^{(\ell)}\right)$ and $\tilde{W}_r \sim q\left(W_{ij}^{(\ell)}\right)$. We can then optimise our lower bound with respect to the parameters governing our variational distributions $\left(m_{ij}^{(\ell)}, (s^2)_{ij}^{(\ell)}, \mu_{ij}^{(\ell)} \text{ and } (\beta^2)_{ij}^{(\ell)}\right)$ and our kernel hyperparameters $\bm{\Theta}$, using conventional gradient descent techniques. The variational lower bound on the marginal likelihood can be derived as follows,
\begin{align*}
\log[p(\bm{y} \mid \bm{X}, \bm{\Theta})] &= \log \left[\int p(\bm{y} \mid \bm{X}, \bm{\Psi}, \bm{\Theta}) p(\bm{\Psi} \mid \bm{\Theta}) d\bm{\Psi} \right] \\ 
&= \log \left[\int \frac{p(\bm{y} \mid \bm{X}, \bm{\Psi}, \bm{\Theta}) p(\bm{\Psi} \mid \bm{\Theta})}{q(\bm{\Psi})} q(\bm{\Psi}) d\bm{\Psi} \right] \\
&= \log \left[E_{q(\bm{\Psi})} \frac{p(\bm{y} \mid \bm{X}, \bm{\Psi}, \bm{\Theta}) p(\bm{\Psi} \mid \bm{\Theta})}{q(\bm{\Psi})} \right] \\
&\geq E_{q(\bm{\Psi})} \left( \log \left[ p(\bm{y} \mid \bm{X}, \bm{\Psi}, \bm{\Theta}) \right] \right) + E_{q(\bm{\Psi})} \left(\log \left[\frac{p(\bm{\Psi} \mid \bm{\Theta})}{q(\bm{\Psi})}\right] \right) \\
&= E_{q(\bm{\Psi})} \left( \log[p(\bm{y} \mid \bm{X, \Psi, \Theta})] \right) - \text{D}_\text{KL}[q(\bm{\Psi}) \mid \mid p(\bm{\Psi} \mid \bm{\Theta})] \\
&\approx \left[\frac{N}{M} \sum_{k \in \mathcal{I}_M} \frac{1}{N_{\text{MC}}} \sum^{N_{\text{MC}}}_{r=1} \log[p(\bm{y}_k \mid \bm{x}_k, \tilde{\bm{\Psi}}_r, \bm{\Theta})]\right] - \text{D}_{\text{KL}}[q(\bm{\Psi})||p(\bm{\Psi} \mid \bm{\Theta})] .
\end{align*}

\subsection{DLFM-VIP}
We can write down our variational lower bound as,
\begin{align*}
    \mathcal{L} &= \mathbb{E}_{q(\{\mathbf{u}^\ell, \mathbf{V}^{\ell} \}^L_{\ell = 1})} \left[\frac{p(\mathbf{Y}, \{\mathbf{u}^\ell, \mathbf{V}^{\ell} \}^L_{\ell = 1})}{q(\{\mathbf{u}^\ell, \mathbf{V}^{\ell} \}^L_{\ell = 1})} \right] = \int q(\{\mathbf{u}^\ell, \mathbf{V}^{\ell} \}^L_{\ell = 1}) \log \left[\frac{p(\mathbf{Y}, \{\mathbf{u}^\ell, \mathbf{V}^{\ell} \}^L_{\ell = 1})}{q(\{\mathbf{u}^\ell, \mathbf{V}^{\ell} \}^L_{\ell = 1})} \right] d\mathbf{S} ,
\end{align*}
where $d\mathbf{S}$ represents the integral over all of the inducing points and latent processes, across all layers of the model. As discussed in the main text, we can write the joint distribution of the DLFM-VIP as,
\begin{align*}
    p(\mathbf{Y}, \{\mathbf{u}^\ell, \mathbf{V}^{\ell} \}^L_{\ell = 1}) =\prod^{N}_{i=1}  p(\mathbf{y}_{i} \mid \mathbf{F}_i^L)
\prod^{L}_{\ell=1} p(\mathbf{u}^\ell \mid \mathbf{V}^{\ell}) p(\mathbf{V}^{\ell}) ,
\end{align*}
and we construct our variational posterior as follows,
\begin{align*}
    q(\{\mathbf{u}^\ell, \mathbf{V}^{\ell} \}^L_{\ell = 1}) = \prod^L_{\ell = 1} p(\mathbf{u}^\ell \mid \mathbf{V}^{\ell})q(\mathbf{V}^{\ell}).
\end{align*}
By substituting these two expressions into our variational lower bound, we can derive a form of the bound which we can use to perform approximate inference as follows:
\begin{align*}
    \begin{split}
        \mathcal{L} &= \int \prod^L_{\ell = 1} p(\mathbf{u}^\ell \mid \mathbf{V}^{\ell})q(\mathbf{V}^{\ell})  \log \left[\frac{\prod^{N}_{i=1}  p(\mathbf{y}_{i} \mid \mathbf{F}_i^L)
        \prod^{L}_{\ell=1} \cancel{p(\mathbf{u}^\ell \mid \mathbf{V}^{\ell})} p(\mathbf{V}^{\ell})}{\prod^L_{\ell = 1} \cancel{p(\mathbf{u}^\ell \mid \mathbf{V}^{\ell})}q(\mathbf{V}^{\ell})} \right] d\mathbf{S}
    \end{split} \\
    \begin{split}
        &= \int \prod^L_{\ell = 1} p(\mathbf{u}^\ell \mid \mathbf{V}^{\ell})q(\mathbf{V}^{\ell}) \log \left[\frac{\prod^{N}_{i=1}  p(\mathbf{y}_{i} \mid \mathbf{F}_i^L)
        \prod^{L}_{\ell=1} p(\mathbf{V}^{\ell})}{\prod^L_{\ell = 1} q(\mathbf{V}^{\ell})} \right] d\mathbf{S}
    \end{split} \\
    \begin{split}
        &= \int \prod^L_{\ell = 1} p(\mathbf{u}^\ell \mid \mathbf{V}^{\ell})q(\mathbf{V}^{\ell}) \log \left[\prod^{N}_{i=1}  p(\mathbf{y}_{i} \mid \mathbf{F}_i^L)\right] d\mathbf{S} - \sum^L_{\ell = 1} \text{D}_\text{KL}\left[q(\mathbf{V}^{\ell}) \lVert p(\mathbf{V}^{\ell})\right]
    \end{split} \\
    \begin{split}
        &= \sum^N_{i=1} \mathbb{E}_{q(\{\mathbf{u}^\ell, \mathbf{V}^{\ell} \}^L_{\ell = 1})} \left[\log p(\mathbf{y}_{i} \mid \mathbf{F}_i^L) \right] - \sum^L_{\ell = 1} \text{D}_\text{KL} \left[q(\mathbf{V}^{\ell}) \lVert p(\mathbf{V}^{\ell})\right]
    \end{split}
\end{align*}
As mentioned previously, the KL terms have a well known tractable form, however we must approximate the intractable expectation term stochastically using $S$ Monte Carlo samples,
\begin{equation*}
    \mathbb{E}_{q(\{\mathbf{u}^\ell, \mathbf{V}^{\ell} \}^L_{\ell = 1})} \left[\log p(\mathbf{y}_{i} \mid \mathbf{F}_i^L) \right] \approx \frac{1}{S} \sum^S_{s=1} \log p\left(\mathbf{y}_i \mid \mathbf{F}_i^{L^{(s)}}\right).
\end{equation*}
where $\mathbf{F}_i^{L^{(s)}}$ is a sample from the model.

\section{Experimental Details}
The experiments performed in this work were primarily carried out using HPC nodes with 80GB NVIDIA A100-SXM4 GPUs, alongside some 16GB Tesla V100-SXM2 GPUs which were used to run some of the less computationally intensive baselines.
We initialise the likelihood variance in all models tested to 0.01, and use a learning rate of 0.01 for the optimiser. 

\textbf{RFF-based models:} In the DLFM-RFF and DGP-RFF, we concatenate the output of the inner layers with the original input to the model as a means of avoiding pathological behaviour \citep{duvenaud2014avoiding}. For both of these models we use 100 random features and 100 Monte Carlo samples across all experiments, initialising all lengthscales (and decay parameters in the DLFM-RFF) to 0.01 unless otherwise specified. The sensitivity parameters in the DLFM-RFF are randomly initialised using a standard normal distribution.

\textbf{Inducing points-based models:}
For all inducing-points based models tested, we use 100 inducing points (per layer), and initialise the inducing inputs at each layer using $k$-means clustering. Time series problems are the exception to this (e.g. the toy and PhysioNet experiments), where at the first layer of the DLFM-VIP we use fixed, regularly-spaced inducing inputs, an approach which has been shown to work well empirically for time series problems \citep{bui2014tree}. 

In order to avoid the aforementioned pathological behaviour common in DGPs, in the case of the DLFM-VIP and DGP-DSVI, rather than feeding forward the original inputs, we instead take the alternative approach proposed by \citet{salimbeni2017doubly}, which entails using a linear mean function at each layer. An additional point to note regarding the DGP-DSVI is that we attempted to use more Monte Carlo samples in order to obtain parity with the number used by our model (particularly the DLFM-VIP), however the best results we achieved overall with the DGP-DSVI were obtained with just a single Monte Carlo sample, as in the original implementation.

\subsection{Toy experiment}
The data was generated by solving the hierarchical ODE system with $\gamma_1 = 0.01$, $\gamma_2 = 0.02$, $\omega=1$, $\tau=0$ and the initial values of $f_1$ and $f_2$ set equal to zero. The noise corruptions for the output were generated from a normal distribution, such that $\epsilon \sim \mathcal{N}(0, 0.04)$. The extrapolation test data consists of 150 evenly spaced data-points between $t=1.0$ and $t=1.25$, whilst the imputation test data consists of 100 evenly spaced data-points between approximately $t=0.208$ and $t=0.375$.

All three deep models tested used a single hidden layer with a dimensionality of 3, and a minibatch size equal to the size of the training data. For the DLFM-RFF and DLFM-VIP, a single latent force was used.

\subsection{PhysioNet Multivariate Time Series}
For the extrapolation experiment, the training data consisted of the 975 data-points lying between $t=0$ and $t=0.975$, whilst the test data consisted of the 25 data-points lying between $t=0.975$ and $t=1.0$. For the imputation experiment, the test data consisted of 150 contiguous, non-overlapping data-points from each of the three outputs; these were in the range $t = 0.20$ to $t = 0.35$ for ABP, $t = 0.40$ to $t = 0.55$ for ECG and $t = 0.60$ to $t = 0.75$ for ICP. For both models, across both experiments, we used two layer models with a hidden layer dimensionality $D_{F^{(\ell)}}=3$, and a single latent force ($Q = 1$).

For the DLFM-RFF, we use 1 Monte Carlo sample for the first 100 iterations before increasing this to the full complement of 100. We also fixed the variational parameters for the first 200 iterations, and the kernel hyperparameters for the first 300 iterations, in order to stabilise training. For the DLFM-VIP, we use 50 Monte Carlo samples from the outset, initialising all lengthscales to 0.1, and all decay parameters to 2.5.

A point to note regarding the DGP-DSVI for the PhysioNet experiment (as well as the toy problem), is that rather than using the original DGP-DSVI implementation as we did for the UCI experiments, we used a more recent implementation of the DGP-DSVI which utilises GPyTorch \citep{gardner2018gpytorch}. The models are broadly similar, the only minor differences are that the GPyTorch model uses 10 Monte Carlo samples rather than a single sample, and it also employs conjugate gradient methods to accelerate convergence.

\subsection{UCI regression}
The \textit{power} and \textit{protein} datasets we use are freely available from the UCI Machine Learning Repository \citep{Dua2019}. For these regression experiments, 90\% of each dataset is used for training and the remaining 10\% is withheld in order to compute the test set metrics presented in the main text. 

For initialisation and training of the DGP-RFF, we follow the procedure used by \citet{cutajar2017random} as closely as possible. We initialise all lengthscales to $\log(D_{F^{(\ell)}})$, where $D_{F^{(\ell)}}$ is the dimensionality of the $\ell$-th layer; for the hidden layer, we set $D_{F^{(\ell)}}=3$. We train until convergence, for 20,000 iterations using a minibatch size of 200. For the first 10,000 iterations we use a single Monte Carlo sample before increasing this to the full complement of 100. We also fix the kernel hyperparameters, denoted by $\boldsymbol{\Theta}$, for the first 12,000 iterations to stabilise training.

We mirror the DGP-RFF settings very closely for the DLFM-RFF, with the only differences being that we share the 100 random features across 5 latent forces (i.e. 20 features per latent force) and we train for 100,000 iterations in order to reach convergence. The settings for the DLFM-VIP broadly mirror those used for the PhysioNet experiment, except that we train for 100,000 iterations with a batch size equal to the size of the dataset for \textit{power}, whilst we train for 50,000 iterations with a batch size of 5000 for \textit{protein}.

\section{Computational complexities}
The computational complexity associated with performing inference in the DLFM-RFF is $\mathcal{O}(m D Q N_{RF} N_{MC})$ where $m$ denotes the mini-batch size, $D$ denotes the layer dimensionality, $Q$ and $N_{RF}$ represent the number of latent forces and random Fourier features respectively and $N_{MC}$ denotes the number of Monte Carlo samples used. This very closely follows the computational complexity of the DGP with random feature expansions presented by \citet{cutajar2017random}, with an added linear dependency on $Q$, the number of latent forces employed.

Inference in the DLFM-VIP is associated with a complexity of $\mathcal{O}(QM^3)$, where again $Q$ denotes the number of latent forces, and $M$ the number of inducing points.

\vskip 0.2in
\bibliography{references}

\end{document}